\documentclass[twoside,11pt]{article}

\usepackage{blindtext}

\usepackage[preprint]{jmlr2e}
\usepackage{amsmath}
\usepackage[T1]{fontenc}
\usepackage[utf8]{inputenc}
\usepackage[french]{babel}
\usepackage{hyperref}
\hypersetup{colorlinks,citecolor=blue,urlcolor=blue,linkcolor=red}
\usepackage{siunitx}

\usepackage{lastpage}
\jmlrheading{23}{2022}{1-\pageref{LastPage}}{1/21; Revised 5/22}{9/22}{21-0000}{Mukherjee and Chang}

\ShortHeadings{Hilbert Space Embedding of Categorical Variables}{Mukherjee and Chang}
\firstpageno{1}

\begin{document}

\title{Addressing Dynamic and Sparse Qualitative Data: A Hilbert Space Embedding of Categorical Variables}

\author{\name Anirban Mukherjee \email am253@cornell.edu \\
       \addr Samuel Curtis Johnson Graduate School of Management\\
       Cornell University\\
       Ithaca, NY 14850, USA
       \AND
       \name Hannah H. Chang \email hannahchang@smu.edu.sg \\
       \addr Lee Kong Chian School of Business\\
       Singapore Management University\\
       Singapore 178899}

\editor{My editor}

\maketitle

\begin{abstract}%   <- trailing '%' for backward compatibility of .sty file

We propose a novel framework for incorporating qualitative data into quantitative models for causal estimation. Previous methods use categorical variables derived from qualitative data to build quantitative models. However, this approach can lead to data-sparse categories and yield inconsistent (asymptotically biased) and imprecise (finite sample biased) estimates if the qualitative information is dynamic and intricate. 

We use functional analysis to create a more nuanced and flexible framework. We embed the observed categories into a latent Baire space and introduce a continuous linear map---a Hilbert space embedding---from the Baire space of categories to a Reproducing Kernel Hilbert Space (RKHS) of representation functions. Through the Riesz representation theorem, we establish that the canonical treatment of categorical variables in causal models can be transformed into an identified structure in the RKHS. Transfer learning acts as a catalyst to streamline estimation---embeddings from traditional models are paired with the kernel trick to form the Hilbert space embedding.

We validate our model through comprehensive simulation evidence and demonstrate its relevance in a real-world study that contrasts theoretical predictions from economics and psychology in an e-commerce marketplace. The results confirm the superior performance of our model, particularly in scenarios where qualitative information is nuanced and complex.

\end{abstract}


\section{Introduction}

Qualitative data refers to non-numerical information used in data analysis. It encompasses attributes or characteristics such as color, taste, texture, smell, or appearance. Derived from interviews, focus groups, observations, and open-ended survey responses, this descriptive data offers deep insights into behaviors, emotions, experiences, and social phenomena.

Integrating qualitative data into quantitative models for causal estimation presents challenges. A prevalent strategy involves transforming qualitative data into categorical variables for inclusion in quantitative models \citep{powers2008statistical}. However, this method can lead to estimates that are both imprecise and biased in a finite sample (i.e., finite-sample bias) or inconsistent, exhibiting bias that does not vanish as the sample size increases (i.e., asymptotic bias) when handling dynamic and complex qualitative information.

Specifically, dynamic information environments exhibit swift and continuous evolution. For instance, on e-commerce platforms like Amazon, new products with unique qualitative descriptors (e.g., shape, style, and look) are released. Concurrently, older products with other qualitative descriptors become obsolete. Similar scenarios occur in healthcare research, where new diseases and treatments emerge \citep{murray2012disability}, network science research with changing node interactions \citep{holme2012temporal}, and climate science research facing continuously shifting climate patterns \citep{knutti2010challenges}. In each of these instances, a portion of the description (e.g., the symptoms of a disease) is communicated qualitatively.

In these examples, as the qualitative data changes over time, so do the categorical variables used to express the qualitative data, and this dynamic can significantly impact data analysis. When categories fade or become obsolete, the accumulation of relevant information ceases. For example, in a demand model, additional observations may relate to new products because outdated products have left the market, limiting the sample suitable for estimating parameters tied to obsolete products (e.g., brand fixed effects, coefficients relating to older technologies and fashion elements) \citep{broda2010product, mukherjee2011modeling}. This can yield imprecise and inconsistent estimates because only fixed-length partitions of the samples are pertinent for estimating corresponding sets of parameters; the information in a period is neither relevant to outdated products nor to future products.

Moreover, in scenarios requiring intricate categorization, some categories may correspond to only a few observations. Consider, for instance, Netflix's in-house movie classification system, which comprises 76,000 microgenres such as `Cult Evil Kid Horror Movies,' `British set in Europe Sci-Fi \& Fantasy from the 1960s,' and `Time Travel Movies starring William Hartnell' \citep{madrigal2014}. The system's complexity likely results in sparse categories in a given dataset, as such specific category definitions can only correspond to a few movies, and thus, to a limited number of observations. Consequently, the estimates of coefficients associated with such sparse categories may exhibit severe finite-sample bias \citep{kim2006blockwise, meier2008group}.

The number of observations within a category may not necessarily correspond to its informational significance. For instance, in a wine dataset, the coefficient associated with the taste descriptors associated with rare but expensive wines could hold substantial informational value. However, conventional approaches aimed at managing data and model complexity may either: merge the category describing the descriptors with other categories, thereby rendering all related category coefficients equivalent, or assume it to be zero by eliminating the corresponding dummy variable. As such, in many social science applications, including this example, the qualitative data components represented by sparse categories are likely to exert a non-zero and distinct influence on the outcome. Therefore, accurately estimating coefficients on data-sparse categories may be pivotal to the analysis.

Lastly, imprecision and inconsistency in category-level fixed effects, or `group-level' parameters, can cause inconsistency in the estimates of the structural parameters of the model (i.e., parameters relating to the entire dataset), a phenomenon known as the incidental parameters problem \citep{fernandez2009fixed, greene2004behaviour, honore2000panel, lancaster2000incidental, manski1987semiparametric, small1993nonparametric}. Furthermore, the exclusion of relevant variables to minimize variance in the bias-variance tradeoff, guided either by domain-specific information or principled variable selection methods such as the Least Absolute Shrinkage and Selection Operator (LASSO; \citealp{chetverikov2021cross, hastie2015statistical, tibshirani1996regression, yuan2006model}), has been shown to significantly bias estimates of the included variables in the model \citep{wuthrich2021omitted}. These factors constrain the classical model's ability to accurately depict the true underlying relationships in the data-generating process.

\subsection{Color Mentions in Amazon Product Titles: A Case Study}
\label{color_Amazon_initial}

\begin{figure}[tbp]
\centering
\includegraphics[width=\linewidth]{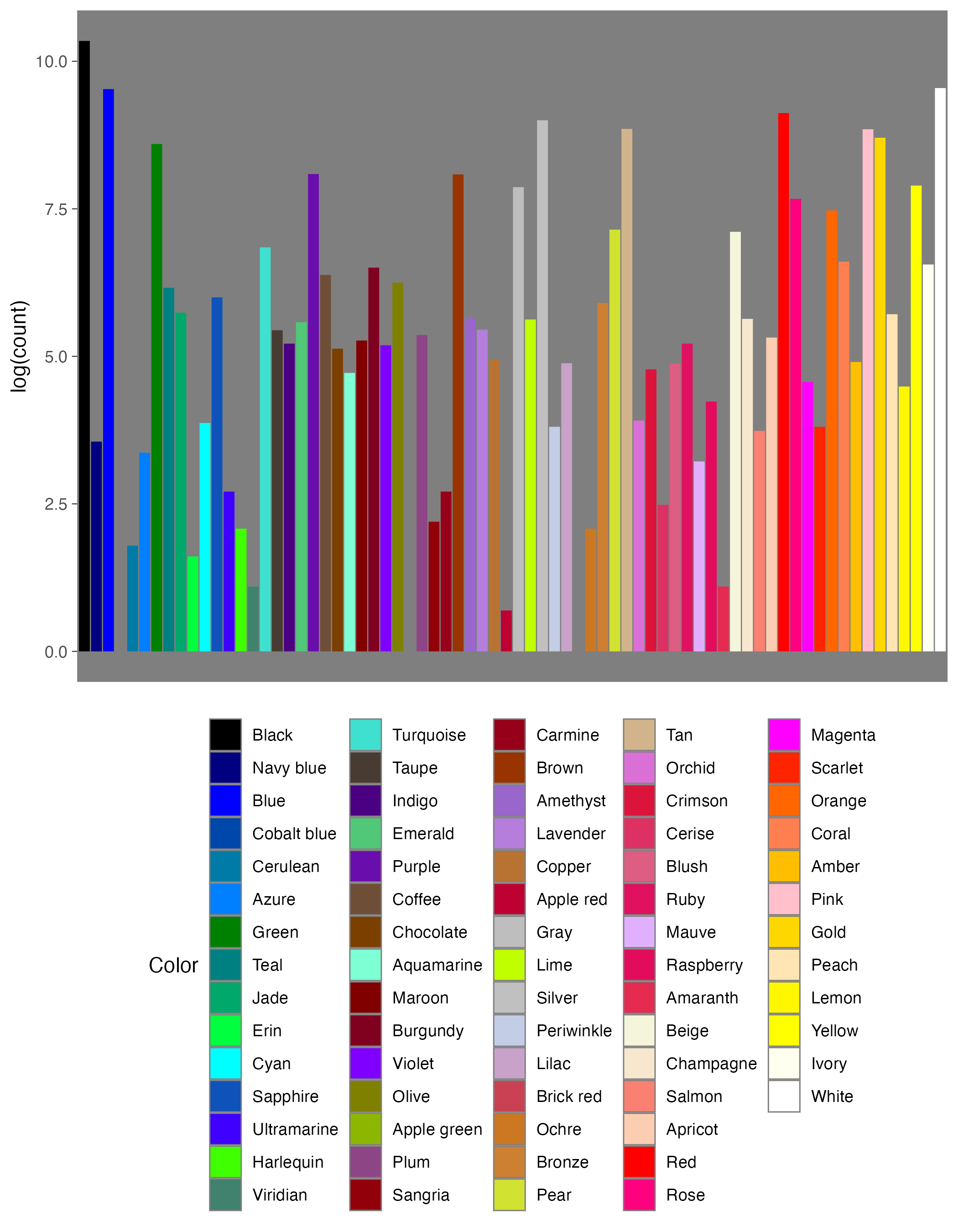}
\caption{Frequency Analysis of Color References in Product Titles}
\begin{minipage}{\linewidth}
\medskip
\footnotesize
Note: The length of each bar represents the natural logarithm of the number of times each color was referenced in a product title. The colors are organized by their RGB representation, with red, green, and blue channels increasing from left to right.
\end{minipage}
\label{fig:colors}
\end{figure}

As a concrete example of the dynamism and complexity of qualitative information, consider color references on Amazon. Figure \ref{fig:colors} presents a frequency analysis of color mentions in product titles in the `Fashion' category. To fix ideas, we consistently refer back to this example throughout our paper.

To construct this figure, we began with a large-scale dataset of Amazon metadata, as detailed by \cite{ni2019justifying}, focusing exclusively on fashion products. We then searched these product titles for a selection of 90 color names that were chosen based on the presence of a corresponding Wikipedia article, to ensure the names are in common usage and are likely to influence consumer behavior. Our final data includes 105,275 products.

The bars in the figure represent the number of product titles that reference each color, presented on a logarithmic scale to compensate for the highly skewed nature of the data. For example, `Black' is the most commonly referenced color, appearing in 31,103 instances, while the 25\textsuperscript{th} and 75\textsuperscript{th} percentiles account for 3 and 506.25 references, respectively. Despite our search being confined to common color names and solely including products within the `Fashion' category, we identified 72 unique colors, demonstrating a wide spectrum of references. The count of unique colors would increase if our search were to encompass less common names or a broader range of product types.

The data comprises a variety of both similar (`Lemon' and `Lime') and dissimilar (`Lemon' and `Raspberry') color names. Suppose we were to specify a demand model based on this data. Traditional categorical variable approaches would treat these colors as entirely unrelated, thereby failing to capture the latent relationships. Crucially, these latent relationships, while not evident in a simple categorical variable for color, are readily discernible in color composition models such as the RGB model. For instance, the similarities between the colors `Lemon' and `Lime', as well as the differences between `Lemon' and `Raspberry', are well-reflected in their respective RGB values. At the core of our proposed modeling framework is the utilization of such latent similarities in the information represented by these categories. This can be achieved either by applying an alternate model to the category labels or by analyzing the physical phenomena they represent (for example, examining the actual colors corresponding to the color names) to enhance estimation efficiency.

\subsection{Overview of Contributions}

This paper introduces a novel machine learning and econometrics framework for integrating qualitative data into quantitative models for causal estimation. Our approach addresses the key challenges of data sparsity, evolving categorizations, and complex categorical structures, thereby improving upon standard models. The primary contributions include:

\begin{enumerate}

\item \textbf{Advanced Processing of Qualitative Data:} We use functional analysis to map observed categories to a latent Baire space, which we then embed in a Reproducing Kernel Hilbert Space (RKHS). This approach offers more nuanced handling of categorical variables than traditional methods.

\item \textbf{Addressing Inconsistency and Imprecision in Dynamic Environments:} Our framework provides consistent estimators in dynamic settings, even where traditional methods may fail. It efficiently handles categories that may become obsolete due to rapid environmental changes.

\item \textbf{Handling Data-Sparse Categories:} Our estimator is designed for complex categorization systems with sparse data. We use Hilbert space embeddings to enhance the interpretation and representation of this data, which increases econometric estimation efficiency.

\item \textbf{Incorporating Supplementary Data:} Our study illustrates how auxiliary information can be used in the modeling process, such as physical phenomena models related to categories (e.g., color representation models), inferences from language models based on textual category descriptions, and transformations of direct stimuli.

\item \textbf{Demonstration of Model Efficacy and Traditional Model Limitations:} We validate our model through rigorous simulation experiments, establishing superior performance in handling dynamic and complex qualitative information and uncovering traditional models' limitations.

\item \textbf{Balancing Intuition and Mathematical Admissibility:} Our research elucidates the following dichotomy between intuition and mathematical admissibility---certain intuitive functions of the initial embeddings such as the dot product are admissible under Mercer's theorem, while others (e.g., the Euclidean norm) are inadmissible. In developing theory, we provide a rigorous foundation that is flexible in offering alternatives, while clearly outlining key requirements.

\end{enumerate}

By applying our model to a novel research question relating to color references in Amazon product titles, we showcase how it can address key challenges and integrate qualitative data into quantitative causal models.

\section{Conceptual Framework}

In this section, we establish the conceptual framework for our research. We begin with a review of the relevant statistical, econometric, and empirical modeling literature. Next, we outline the mathematical structures that underpin our model. We conclude with a discussion of our analytical and empirical studies, and how they relate to the extant literature.

\subsection{Prior Literature}

Despite the pervasiveness of qualitative data and the challenges posed by the sparse and dynamic categorical variables it generates, comprehensive solutions have yet to be found. Practitioners typically employ one of two strategies: either consolidating rare categories or implementing variable selection methods \citep{hastie2015statistical, tibshirani1996regression, yuan2006model}. However, these strategies fall short in several aspects. They fail to provide estimates for rarely observed or emerging categories, often leading to these being dropped or consolidated. Furthermore, they struggle to address potential endogeneity issues, which can arise from the loss of unobserved qualitative information when categories are merged or dropped. Specifically, while variable selection models like LASSO offer consistent estimation in sparse variable scenarios where many coefficients are nonzero \citep{zou2006adaptive}, they introduce bias in situations with dense relationships, where many or all coefficients are nonzero---a manifestation of the bias-variance tradeoff \citep{vapnik1999nature, wuthrich2021omitted}.

Existing methods are partly limited by the approach of considering high-dimensional categorical parameters as `nuisance' parameters, with an assumption that they are not the central point of investigation. This perspective is evident in the Neyman-Scott paradox. In this paradox, the group means (nuisance parameters) are estimated with consistency but not precision. Conversely, the standard deviation of these group means (the structural parameter) is estimated precisely but inconsistently. Current methods mainly aim to restore consistency in the structural parameter estimates at the expense of the nuisance parameter estimates \citep{greene2004behaviour, honore2000panel, small1993nonparametric}. For instance, the analogue of the Neyman-Scott paradox in panel data is typically addressed by differencing out the group-level panel fixed-effects. While this method can produce consistent estimates of the structural parameters, it does not provide estimates of the panel-level nuisance parameters \citep{lancaster2000incidental, manski1987semiparametric}.

With the proliferation of data sources and the advancement of large-scale databases cataloging human activity, it is increasingly common for economic models to be situated within high dimensional data. Consequently, there's a growing appreciation for sparse or approximately sparse models, where numerous variables and their transformations are candidates for inclusion in a statistical model but only a few are principal, resulting in a low-rank true design matrix (e.g., \citealp {athey2018approximate, belloni2011high, bickel2009simultaneous, chen2022fast, chernozhukov2021inference, d2021overlap, fan2023latent, farrell2015robust, vanderweele2019principles, zhang2008sparsity}). Alternatively, in many cases, the empirical process can be well approximated by a sparse model or a model with regularized coefficients, such as those proposed by \cite{belloni2014high} and \cite{candes2005dantzig}.

To handle high parametric complexity and high dimensionality, contemporary econometrics research methods often turn to machine learning. For instance, in the treatment effects literature, it is increasingly common to initially estimate two machine learning models---one for the outcome and the other for the propensity score---followed by the estimation of a causal econometric model \citep{chernozhukov2018double}. In such instances, the causal estimator can significantly benefit from techniques such as sample splitting, cross-fitting, and specifying Neyman orthogonal moment conditions \citep{chernozhukov2017double}.

Distinct from these use-cases, our research focuses on applications where complex and dynamic categorical information holds primary importance. For instance, researchers may wish to include a brand fixed effect in an Amazon demand model or a micro-genre-fixed effect in a Netflix demand model. In such situations, estimating the fixed effects of data-sparse categories, i.e., categories with corresponding observations, can be crucial. Take, for instance, the challenge of estimating the fixed effects of watch micro-brands like Autodromo and Farer in a demand-side analysis. Estimating these parameters may be challenging due to their infrequent appearance in the data. However, these estimates may be indispensable for a comprehensive analysis if the research question pertains to brands and branding such that micro-brands play a significant role. Although these types of scenarios are commonplace, they have received comparatively less attention in the field.

\begin{figure}[tbp]
\centering
\includegraphics[width=0.81\linewidth]{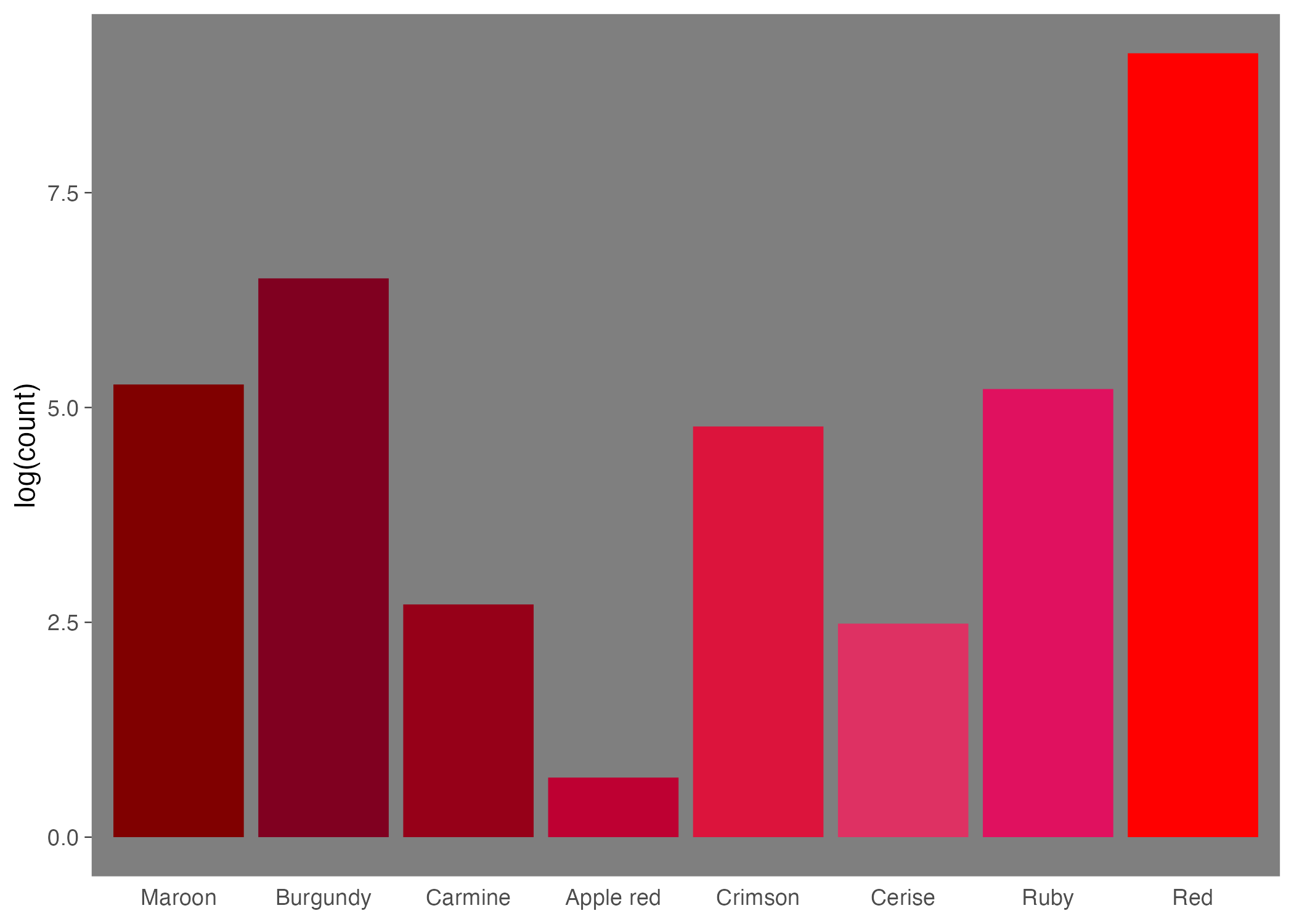}
\caption{Frequency Analysis of Shades of Red in Amazon Product Titles}
\begin{minipage}{\linewidth}
\medskip
\footnotesize
Note: The length of each bar represents the natural logarithm of the number of times each color was referenced in fashion product titles on Amazon. Eight representative shades are organized by their RGB representation.
\end{minipage}
\label{fig:colors_red}
\end{figure}

We accomplish these goals by identifying and incorporating categories based on commonalities found within the existing data. For example, even if products in `Carmine' and `Cerise' are infrequent, our model can infer the fixed effects for these colors by analyzing data related to other shades of red, such as `Maroon', `Burgundy', and `Red'. Figure \ref{fig:colors_red} shows the frequency of fashion product titles on Amazon referencing eight popular shades of red. This visual representation underscores the intuition behind our model, demonstrating how color categories, although closely related, might be referenced with varying frequencies due to linguistic conventions. Such variations lead to sparse categorical data, which our method navigates by learning and leveraging inter-category relationships.

Modeling data-sparse categories carries important implications for the research process. The results generated by existing qualitative data management strategies often hinge on the specifics of a particular dataset. This is because the inclusion and exclusion of categories, and thus any bias arising from omitting relevant variables, depend on the frequency of these variables in a specific dataset, rather than their theoretical significance.

Recall our earlier example of Netflix's micro-genre categorization of movie storylines. If we were to conduct a demand analysis for Netflix, one data complication that may arise is that the prevalence of micro-genres in a dataset is likely to fluctuate over time. This fluctuation occurs because movies often evolve in cycles or waves, with themes and storylines recurring over time. Various factors, including societal shifts, technological advances, and commercial successes that inspire imitation, contribute to these trends \citep{mukherjee2018}. For instance, the success of films like `The Matrix' spurred an influx of technologically-driven, dystopian narratives at the turn of the 21\textsuperscript{st} century \citep{neale2005genre}.

If the inclusion of categories is determined by the frequency with which the categories appear in the data, then the micro-genres included in the analysis depend on the data selection criteria (e.g., which years are covered by the data and which movies are included in the data). Thus, for example, a study of movies released prior to the release of `The Matrix' would be based on data that reflects a different frequency of micro-genres, and therefore would employ a different set of micro-genres in the analysis, than a study of movies released after `The Matrix'.

This scenario underscores a broader problem in managing qualitative data: the absence of a standardized approach to category selection complicates the comparison of findings across studies, exacerbating the challenge of managing sparse and dynamic categorical variables. Our method alleviates this issue by automatically projecting categorizations into a continuous space. As this process applies the same mapping, regardless of covariate balance, across all studies, it enables researchers to compare findings in a more meaningful way.

In sum, our research addresses data sparsity in dense statistical models (i.e., where many or all of the coefficients on variables are nonzero). In classical settings, where the size of the category set and its prevalence in the data remain constant over time (i.e., categories do not become obsolete), estimating fixed effects is straightforward. However, in data-sparse settings, excluding fixed effects is likely to bias the inferences and reduce their specificity, whereas including them is likely to lead to imprecision and therefore the incidental parameters problem. Therefore, our research is driven by different considerations than those of the existing literature; our focus arises from the complexity of qualitative data, as represented in a glut of categorical variables, rather than from a large number of quantitative variables. Our proposed methodology aims to bridge this critical gap.

\subsection{Overview of Our Approach}

We introduce a model that integrates principles from econometrics, functional analysis, and machine learning to adeptly handle categorical variables. This subsection offers a high-level outline of the model's principal components, with subsequent sections providing in-depth explanations.

\begin{enumerate}

\item \textbf{Baire Space of Categories:} Unlike traditional models that treat categories as discrete, orthogonal entities, we embed categories in a latent Baire space. The topology of this space is designed to capture inherent similarities among categories, as reflected in their impact on a target outcome.

\item \textbf{Reproducing Kernel Hilbert Space (RKHS) of Category Representations:} We introduce a latent RKHS of category representation functions. By leveraging the kernel trick in its formulation, we lay the foundation for a comprehensive model of qualitative data, sidestepping the need for its explicit construction.

\item \textbf{Bounded Linear Bijective Operator Between Spaces:} A continuous, linear operator \( T \) bridges the Baire space and the RKHS of representations. Its inverse serves as a continuous linear map from the RKHS back to the category space. This architecture enables us to stipulate a continuous linear functional on the RKHS, which corresponds to the fixed effect linked with each category within the RKHS. This functional symbolizes the conditional mean shift occurring when the reference category is supplanted by its RKHS counterpart.

\item \textbf{Reformulation of the Semiparametric Framework:} By utilizing the defined functional, we augment the semiparametric framework, transitioning from a traditionally discrete structure to a continuous one. This adaptation allows us to seamlessly integrate qualitative data into quantitative models using the RKHS and the kernel trick. As a result, the model's resilience is notably amplified in dynamic situations and when dealing with sparse category-level data.

\item \textbf{Application of the Riesz Representation Theorem:} Using the Riesz Representation Theorem, we establish a one-to-one correspondence between the Riesz representer of the functional and parameters in the original model. This ensures the identifiability of our transformed model.

\item \textbf{Initial Embedding and the Kernel Trick:} We construct an initial embedding \( \Gamma \), mapping the Baire space to a complete inner-product vector space. This mapping is deliberately designed to provide a numerical representation of the data, reflecting its inherent properties, and enabling the use of the kernel trick to compute the inner product in an expanded feature space. This approach enhances computational efficiency while preserving constraints on the dimensionality of the parameter space in the transformed model.

\end{enumerate}

\begin{figure}[tbp]
\centering
\includegraphics[width=\linewidth]{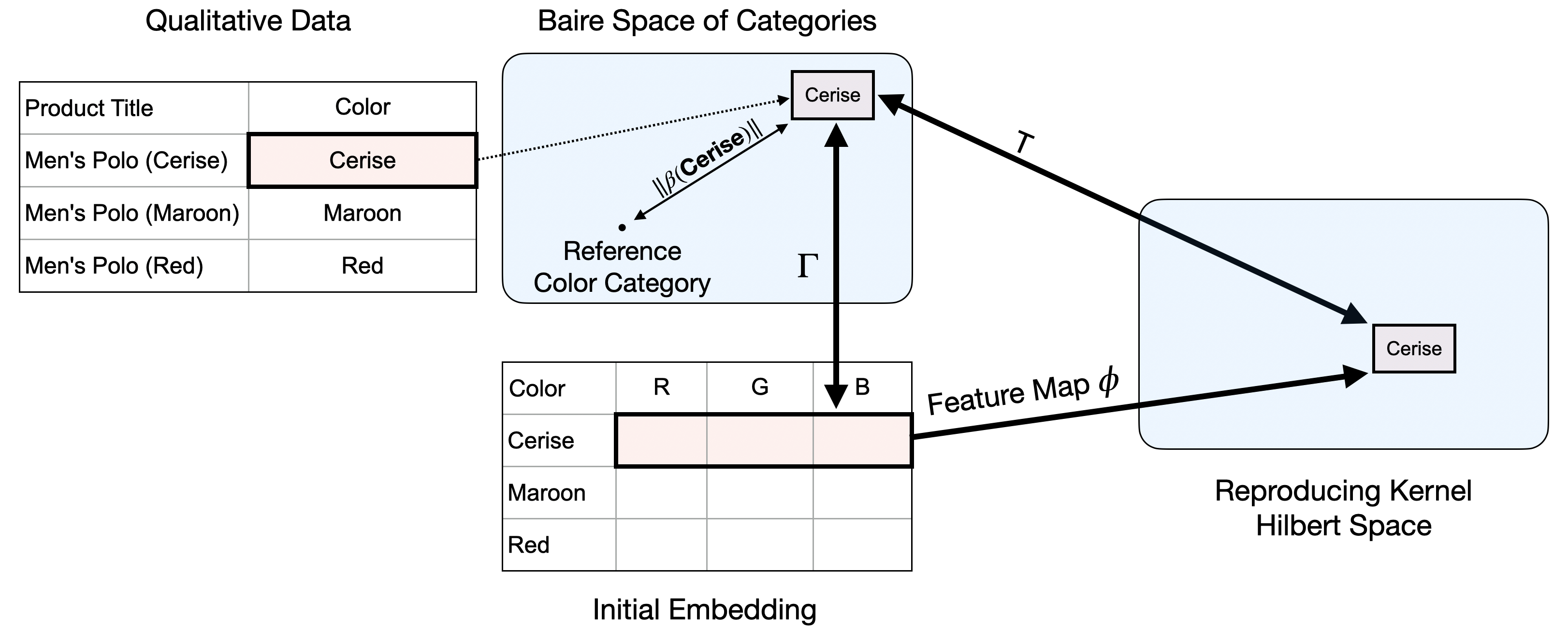}
\caption{An Overview of Our Model}
\begin{minipage}{\linewidth}
\medskip
\footnotesize
Note: This graphic showcases the central components of our model. Qualitative data, represented in tables as descriptions (like color), is channeled into a Baire category space. Distances within this space depict outcome disparities when one category supplants the reference. The map \( \Gamma \) conveys categories to an initial embedding. Operator \( T \) facilitates the connection between the Baire space and an RKHS of representation functions. The feature map \( \phi \), in alignment with the kernel function, enables model operationalization without the necessity of direct computation of features in the high-dimensional RKHS.
\end{minipage}
\label{fig:model_overview}
\end{figure}

Figure \ref{fig:model_overview} provides a visual encapsulation of our innovative approach, emphasizing the synergy of diverse mathematical concepts and computational techniques. By bridging traditional econometric methods with functional analysis and machine learning principles, we present a model capable of adeptly managing categorical variables. This integration promises increased resilience in handling dynamic situations and sparse category-level data, paving the way for new analytical possibilities.

In concert, these elements construct an identifiable empirical process that can be used to compute the conditional mean shift linked to the qualitative data. The following illustrates how qualitative data can be incorporated into a quantitative model on colors. We can form an initial finite-dimensional embedding by applying a color model to the qualitative data. To make an inference about a color (e.g., `Cerise'), we obtain its description in the color model and use our parameter estimates to compute its fixed effect in the econometric model. By construction, the model elements are designed to reflect the conditional mean shift associated with the data represented in the categories. These estimates will then reflect both the information on `Cerise' products and the influence of other colors, varying in their similarity to `Cerise'. For instance, in our data on Amazon products, only 12 are 'Cerise', even though related colors such as `Red' are common (13,992 products in our data). Thus, pooling information across categories (i.e., pooling information across colors based on color similarity) serves a dual purpose: it boosts efficiency and ensures convergence in contexts that do not adhere to classical requirements, such as a bounded number of categories and therefore bounded estimator entropy.

\subsection{Analytical Approach and Empirical Investigation}
\label{sec:AA}

Our research methodology combines a comprehensive analytical approach with an empirical investigation. The analysis employs a reinforcement algorithm based on the seminal work of \cite{yule1925ii} and \cite{simon1955class}. This algorithm generates the Yule-Simon distribution, a discrete probability distribution commonly used to describe the distribution of ranks in various contexts, such as in language and species richness. Named after G. Udny Yule and Herbert A. Simon, this distribution is characterized by a power-law tail, meaning that large ranks can still have substantial probability. Its probability mass function is governed by a parameter often referred to as the reinforcement parameter. 

The Yule-Simon distribution has been employed in various scientific fields, including linguistics, ecology, and economics, particularly in the modeling of processes where new categories can emerge over time or where rich-get-richer dynamics are present. Examples of such phenomena include the number of species per genus in certain higher taxonomic levels of biotic organisms, the size distribution of cities, the distribution of wealth among individuals, and the number of links to webpages on the World Wide Web \citep{garcia2011fixed}. Indeed, this distribution was presented in \cite{simon1955class} as a model for generating text, a prominent form of qualitative data. Importantly, except in the trivial case, the reinforced sequence is neither exchangeable nor partially exchangeable, as we describe later in the paper.

A recent study by \cite{bertoin2020linear} provides several significant insights. First and foremost, although the empirical process is covered by the conclusion of the Glivenko–Cantelli theorem, it is not by the conclusions of the Donsker theorem. Specifically, for certain parameter values, the empirical process converges in law to a Brownian bridge up to a scale factor. For other parameter values, an additional rescaling is necessary, and the limit is a Brownian bridge with exchangeable increments and discontinuous paths.

To illustrate the forms of data we consider, we employ the Yule-Simon distribution to model categories derived from qualitative variables, a subset of the independent variables. Estimating the fixed effects of these categories can become intractable without additional model structure. This issue arises when the number of parameters increases with the number of categories. Consequently, the entropy of the model, as measured by the logarithm of the covering number, increases as a function of \( C_n \). Therefore, if \( C_n \) increases in \( n \), the parameter space may be unbounded in complexity. This is especially evident when examining the empirical measure indexed by \( \mathcal{L} \).

\begin{equation}
\mathcal{P}_i (l) = \frac{1}{i}\sum_{n=1}^i l(\{I_{nc}\}_{c=1}^{C_n}, y_n, v_n; \{\beta_{c}\}_{c=1}^{C_n}, \zeta),\ l \in \mathcal{L},\ y_n \in \mathcal{Y},\ v_n \in \mathcal{V}. \label{eqn:Pi}
\end{equation}

\noindent Here, $l:\mathcal{C} \times \mathcal{Y} \times \mathcal{V} \to \mathbb{R}$ denotes a real-valued loss function that incorporates category fixed effects. $\mathcal{C}$ is the sample space of categories, $\mathcal{Y}$ is the sample space of the dependent variable, and $\mathcal{V}$ is the sample space of other independent variables. $\mathcal{L}$ is a collection of such loss functions. The fixed effect parameter for category $c$ is denoted by $\beta_{c}$. $I_{nc}$ is a dummy variable indicating whether observation $n$ corresponds to category $c$. $C_n$, representing the dependence of the cardinality of the observed categories on $n$, is non-decreasing in $n$. $\zeta$ are the structural parameters.

The empirical measure may fail to converge in distribution to the mean-zero $P$-Brownian bridge. Thus, $l$ is not a $P$-Donsker class function \footnote{A class $\mathcal{F}$ of measurable functions on a probability space $(A, \mathcal{A}, P)$ is called a $P$-Donsker class if the empirical processes $X^P_n(f) \equiv \sqrt{n} (P_n(f) - P(f))$, for all $f \in \mathcal{F}$, converge weakly to a $P$-Brownian bridge with almost surely bounded and uniformly continuous sample paths \citep{sheehy1992uniform}.} if the Yule-Simon model determines the categories denoted by $I_{nc}$  \citep{van1996weak}. Here $P$ is a probability measure.

A failure of the empirical measure to converge occurs if $C_n$ grows with $n$, as previously discussed. However, it is important to note that in this formulation, a failure of the empirical measure to converge may occur even if the category assigment is achieved through a fixed partition of the unit interval that does not change with $n$. In particular, it is not necessary for $C_n$ to increase in $n$. This is clarified by the sequence of indicator functions $\hat{G}_n(u)$, given by
\begin{equation}
\hat{G}_n(u) = \frac{1}{\sqrt{n}}\sum_{i=1}^n{\left( \mathbf{1}_{\hat{U}_i \leq u} - u \right)}, \quad u \in [0, 1]. \label{eqn:IndF}
\end{equation}

This sequence of indicator functions $\hat{G}_n(u)$ defined by Equation \eqref{eqn:IndF} may not converge to the $P$-Brownian bridge if the $\hat{U}_i$ variables are distributed reinforced uniform \citep{bertoin2020linear}. Consequently, the empirical measure in Equation \eqref{eqn:Pi}, in which the loss function incorporates category fixed effects, may not converge in law if the category assignments are defined using the indicator functions $\hat{G}_n(u)$.

We employ the Yule-Simon process due to its wide-ranging applicability. Various refinements have been proposed for other classical processes introducing `innovations' (e.g., the addition of new colors in the Pólya urn process, \citealp{bertoin2022limits, janson2023fluctuations}), `reinforcements' (where, at each step, a process can repeat one of its preceding steps), and `stops' (where, at each step, the process might remain at its current position). A notable example of such enhanced processes is the recently proposed variations of the elephant random walk process \citep{bertenghi2021asymptotic, laulin2022new}. These modifications resonate with real-world contexts where prior categories reappear, new ones surface, and extant ones fade out, thereby aligning the data generating process with many examples of categorical data.

When these features are integrated into statistical processes, they can introduce complex dynamics and disturb the convergence to the Brownian bridge process, an assumption prevalent in many statistical and econometric models. In such contexts, if category incidence indicators are constructed using these models, they might both emulate data generating processes that mirror or align closely with observed qualitative data and present similar challenges due to the growth in parametric entropy. Our model provides a potential alternative: the additional structure we integrate ensures that the parametric complexity remains bounded as the sample size grows, guaranteeing convergence. While our discussion is anchored in the Yule-Simon process, our findings likely hold more widely.

We translate our theoretical insights into practical implications by generating data across varying parameter values using the Yule-Simon reinforcement process. This data is then scrutinized using the standard fixed-effects regression model. Intuitively, one might assume that the conventional model would be robust when the rate of new category emergence is low, especially in qualitative datasets dominated by few recurring elements. This notion originates from the idea that there would be fewer category fixed effects to estimate. However, our results indicate a divergence even in these scenarios.

This divergence can be traced back to the reinforcing nature of the category emergence process. Specifically, categories that appear later in the data sequence are less represented, as the likelihood of selecting an existing category is influenced by its past appearances. This pattern gives rise to a large number of data-sparse categories, leading to an increase in the entropy of the inferred parameter distribution with the addition of new data. This phenomenon persists even in situations where a new category emerges only once in every 100 observations, indicative of a relatively stable information structure.

Next, we apply our model to a widely used e-commerce dataset to explore a theory-driven research question about user ratings for fashion products of different colors. We divide our investigation into two main theoretical perspectives: economic factors and psychological factors.

\textbf{Economic factors}: Economic theory posits that evaluations of products, averaged over a large-scale dataset of potentially functionally identical items that vary only in color, should be equal in distribution. This stems from the notion that any discernible relationship between customer satisfaction and color should be competed away in a market with free entry and exit \citep{dixit1979model, tirole1988theory}. Specifically, if a color is evaluated highly, more products of that color, possibly of lower quality, can be released; conversely, fewer products may be released for colors evaluated poorly. This occurs because fashion products can generally be produced in any color at nearly the same cost. Consider an example where `Blue' shirts are received more favorably than `Black' shirts. Since both can be produced at the same cost, market dynamics involving the entry and exit of products should eliminate this advantage. Formally, in a market characterized by differentiated Bertrand competition with symmetric costs, free entry, and no exit barriers, any advantage provided by a factor that incurs no additional cost, such as color, should not persist \citep{singh1984price}.

\textbf{Psychological factors}: While the economic theory provides a rational perspective, the complexity of human behavior introduces additional variables, as seen in the psychological factors. Reasons stemming from consumer behavior and psychology literature may suggest that products of different colors are received differently by consumers. For instance, prior literature indicates that darker colors might elicit a broader spectrum of emotions, both positive and negative \citep{valdez1994effects}. They are also more attention-grabbing, possibly leading to polarized opinions \citep{gorn1997effects}. The cultural significance of colors, such as black's association with death in Western cultures and with power elsewhere \citep{ho2014colour}, further complicates evaluations. Additionally, personal color preferences may lead to diverse reviews. These preferences are influenced by various factors such as culture, personal experiences, and current trends \citep{palmer2013visual}. Even the perceived quality of a product may be influenced by its color, with colors deemed high-quality or aesthetically pleasing likely to receive more coherent and positive reviews \citep{labrecque2013or}. These factors collectively indicate differential preferences in user reviews and ratings for items of varying colors.

The efficacy of the economic factors hinges on analysts' ability to relate observed outcomes to color references. If analysts fail to infer this relationship, the economic rationale no longer holds, as market participants can only exploit arbitrage opportunities that they can identify. In this scenario, the economic argument may be eclipsed by psychological explanations. Conversely, if the relationship is straightforward to detect, then competitive pressures may dominate.

Crucially, this information is relatively straightforward to infer in the data for frequent colors such as `Blue' and `Black'. However, it may be more complex to discern for infrequent colors such as `Cerise', as previously discussed in Section \ref{color_Amazon_initial}. In such cases, the empirical relationship may be challenging to identify using traditional estimators, causing the economic arguments to falter. Thus, the degree to which a systematic and predictable relationship exists between outcomes and color mentions serves as a barometer of the practical relevance of the estimation issues that motivate our research.

To shed light on this debate, we demonstrate the versatility of our proposed methodology by estimating various specifications of our key innovation---a mechanism to include qualitative descriptors like color in a theoretically grounded causal estimation model. We uncover a clear relationship between color and ratings that is indicative of the influence of the psychological factors over the economic. Moreover, by examining the `Fashion' category on Amazon, we open avenues for practical application and further academic inquiry. This aligns with the intricate intersection of economics, psychology, and consumer behavior that forms the central theme of our investigation.

\section{Model Formulation}

This section lays the groundwork for the mathematical foundations of our proposed framework, focusing on a versatile semiparametric model that strikes a balance between flexibility and interpretability.

We begin by introducing a continuous representation of categories through a Baire space, a mathematical construct that allows for more effective integration of qualitative data. This representation forms the basis for connecting the category space and representation space, achieved through the construction of a RKHS and a bounded linear operator.

Leveraging the Riesz Representation theorem, we reformulate the model to seamlessly incorporate qualitative data while retaining identifiability. We further detail how constructing the representation map through an initial embedding and the kernel trick enhances efficiency. Finally, we touch upon nonparametric specifications and describe the estimation procedure, culminating in a framework adept at handling dynamic and intricate qualitative data.

Our model builds on the semiparametric framework, a key component in achieving the balance between model flexibility and interpretability that we have outlined. Semiparametric models maintain the virtues of parametric models, including having a finite parameter vector that governs the influence of key explanatory variables \citep{li2023nonparametric, robinson1988root, schmalensee1999household}. This aspect facilitates interpretation and inference. At the same time, semiparametric models also incorporate nonparametric components that can model complex, nonlinear relationships without strong modeling assumptions, thereby providing greater model flexibility than pure parametric models. 

In our application, the parametric linear component captures the conditional mean shifts induced by the membership in qualitative categories, our primary explanatory construct of interest. The nonparametric component flexibly models the effects of any additional covariates. This hybrid approach balances interpretability of the effects of interest with model flexibility to accurately characterize the true data generating process. In addition, by restricting the nonparametric complexity to certain model components, our approach retains more estimation efficiency than fully nonparametric models, which can suffer from the curse of dimensionality. We revisit this theme in Section \ref{section:nonparametric}, where we discuss the extension of our model to be fully nonparametric in qualitative information.

With this foundation, our model formulation strives to:

\begin{enumerate}
\item Craft a continuous representation of categories, aiming for deeper insights and relationships, achieved via embedding in a Baire space and then in a RKHS of category representation functions, thereby smoothly integrating qualitative data.
\item Guarantee model identifiability using the Riesz Representation theorem.
\item Boost computational and econometric efficiency through the use of an initial embedding and the kernel trick.
\end{enumerate}

As some target application domains of our model may not regularly involve functional analysis and abstract mathematics, we offer a more detailed description to make our model and paper more broadly accessible. To maintain focus on the novel aspects of our approach and their applications, we reference established theorems and results without delving into the proofs. In general, these proofs can be found in seminal textbooks such as \citealp{berlinet2011reproducing}. Our primary contributions lie in combining insights from these areas of mathematics and statistics with classical formulations of qualitative data estimands and modern embedding methods, to develop novel inference techniques.

\subsection{Semiparametric Framework: An Overview}

We consider semiparametric models of the following form:
\begin{align}
y_n (I_n, v_n) & = \alpha + \sum_{c=1}^{C_n} \beta_c I_{nc} + g(v_n; \zeta) + \epsilon_n, \label{eqn:eqn1a} \\
o_n (I_n, v_n) & = \mathrm{obs}(y_n). \label{eqn:eqn1b}
\end{align}

The latent variable \(y_n\), associated with the \(n^{\text{th}}\) observation, is influenced by two types of variables: the categorical variables derived from qualitative data, represented by binary indicators \(I_n = \{I_{nc}\}_{c=1}^{C_n}\), and the observed quantitative and categorical variables. The indicators denote the membership of the \(n^{\text{th}}\) observation among the \(C_n\) categorical divisions, allowing the model to integrate qualitative data. Importantly, to account for the emergence of new categories and the subsequent expansion of the observed category set, \(C_n\) may increase with respect to \(n\).

The second type, represented by \(v_n \in \mathcal{V}\), incorporates observed quantitative and categorical variables related to the \(n^{\text{th}}\) observation. We define \(\mathcal{V}\) to be a second countable Hausdorff space, aligning it with other semiparametric models. This specification ensures the separation of distinct points and the compactness of the space. \(g(v_n; \zeta)\) denotes a Lipschitz continuous and smooth function, which is parameterized by \(\zeta\) and accounts for the additional nonlinear influences of the covariates \(v_n\) on \(y_n\). These are general conditions that can accommodate many typical empirical specifications.

The model's linear component, \(\alpha + \sum_{c=1}^{C_n} \beta_c I_{nc}\), captures the shift in the conditional mean of \(y_n\) resulting from membership in categories derived from the qualitative data. Each coefficient, \(\beta_c\), represents the change in the conditional mean of \(y_n\) given \(v_n\) when the \(c^{\text{th}}\) category replaces the reference category. For identification purposes, we adopt the regularity conditions that the intercept, \(\alpha\), is synonymous with the conditional mean of the reference category, \(\beta_c = 0\) for the reference category, and \(\beta_c\) is finite for all \(c\).

The error term, \(\epsilon_n\), accounts for the influence of unobserved variables or random fluctuations on the relationship between \(y_n\) and \(o_n\). The observed outcome, \(o_n\), is derived via an observation function, \(\mathrm{obs}\), which links the latent variable \(y_n\) to the observed data. In standard linear regression, \(\mathrm{obs}\) functions as an identity function, thereby directly linking the latent variable \(y_n\) to the observed outcome \(o_n\). However, in more specialized regression models such as censored regression or models for limited dependent variables, \(\mathrm{obs}\) operates as a Càdlàg function. Under certain conditions (e.g., in probit, tobit models), inaccurate estimation of \(\beta\) may bias the estimates of \(\zeta\), even if \(\zeta\) is adequately informed by the data.

Additional restrictions on \(g\) and \(\zeta\) are necessary. For instance, without such restrictions, \(|\zeta| = N\) for \(N\) observations could yield a \(g\) that perfectly fits the data but provides little insight into the true data-generating process. Multiple approaches exist to regularize \(g\). Given our paper's focus on the first type of variables, we abstract from the specific form of restrictions chosen by the researcher except for requiring that Equation \eqref{eqn:eqn1a} is identified conditional on the linear component of the model: \(\zeta\) can be uniquely determined from \(y_n\) given \(\alpha + \sum_{c=1}^{C_n} \beta_c I_{nc}\) and \(v_n\).

We do not require the parameters \(\{\beta_1, \dots, \beta_{C_n}\}\) to be identified in Equation \eqref{eqn:eqn1a}. For example, consider a case where each observation belongs to a category that is `nearly identical' to the preceding one. In this scenario, the classical model is underidentified because \(C_n = n\). The classical model, however, does not use the information that the categories are only `slightly' different between observations. Our model aims to formalize these latent similarities. Therefore, identification in our model is determined in a different parameter space and is not tied to the parameters of the original model. This adjustment ensures the utility of our model, even when the classical model is underidentified.

\subsection{Baire Space of Categories}

To develop a model that more effectively integrates qualitative data into quantitative models, we propose a shift from a discrete to a continuous representation. Specifically, we expand the set of categories to form a latent Baire space (a complete pseudometric space) and denote it as \(\mathcal{X}\). We introduce a continuous, linear function \(\beta: \mathcal{X} \to \mathbb{R}\) that belongs to \(\mathcal{X}\)'s dual space, denoted by \(\mathcal{X}^{\ast}\), and that corresponds to \(\{\beta_1, \dots, \beta_{C_n}\}\) in Equation \eqref{eqn:eqn1a} such that \(x_c\) represents the \(c^{\text{th}}\) category and \(\beta(x_c) = \beta_c\). This function symbolizes the conditional mean shift in \(y\) when a category \(x\) replaces the reference category.

Since \(\beta\) belongs to \(\mathcal{X}^{\ast}\), it is linear, additive, and homogeneous in \(x\). This enhanced capacity of a continuous function better captures the complexity and nuances of qualitative data. Moreover, the specification of a continuous function and associated vector space improves inference, as these model elements account for latent similarities in the categories, as opposed to treating the categories as orthogonal elements in the parameter space.

Together, these modifications enable us to refine the representation \(\sum_{c=1}^{C_n} \beta_c I_{nc}\) in Equation \eqref{eqn:eqn1a} as follows:
\begin{equation}
y_n (x_n, v_n) = \alpha + \int_{\mathcal{X}} \beta(x) \delta(x - x_n) \, dx + g(v_n; \zeta) + \epsilon_n, \label{eqn:eqn2a}
\end{equation}
\noindent where \(\delta(x - x_n)\) refers to the Dirac delta function centered at \(x_n\), which corresponds to the category of the \(n^{\text{th}}\) observation. The Dirac delta function `selects' the effect of the specific category \(x_n\) corresponding to the \(n^{\text{th}}\) observation.

\(\mathcal{X}\) is constructed to conform to Equation \eqref{eqn:eqn1a} and Equation \eqref{eqn:eqn2a} in three key aspects:

\begin{enumerate}
\item We posit that the combination of the latent properties of two categories will correspond to a category with the resultant combined latent properties. Therefore, the addition operation in \(\mathcal{X}\), defined by the union set operation, aligns with the equation:

\begin{equation*}
\beta(x_1 + x_2) = \beta(x^*),
\end{equation*}

\noindent for \(x^*\) whose latent properties are the sum of the latent properties of \(x_1, x_2 \in \mathcal{X}\).
\item We posit that the scalar multiplication of a category's latent properties will correspond to the category possessing the scalar multiple of those latent properties. Therefore, we require that the scalar multiplication operation in \(\mathcal{X}\) aligns with the equation:

\begin{equation*}
\beta(k x_1) = \beta(x^{**}),\ k\in \mathbb{R},
\end{equation*}

\noindent for \(x^{**}\) whose latent properties are \(k\) times the latent properties of \(x_1 \in \mathcal{X}\).
\item The pseudometric on \(\mathcal{X}\) is the Euclidean metric on \(\mathbb{R}\) applied to the distance between the mean shifts induced by the two categories. This metric satisfies the three properties of a pseudometric: it is nonnegative, symmetric, and satisfies the triangle inequality. However, unlike a full metric, this pseudometric may not satisfy the identity of indiscernibles. Even if two categories are distinct within the qualitative data, they could yield the same conditional mean shift, resulting in a distance of zero between them. For example, in a model where qualitative data on colors is mapped to the Baire space, two colors might induce the same conditional mean shift.
\end{enumerate}

\(\mathcal{X}\) is the minimal complete pseudometric space that satisfies the properties outlined above. A Baire space and continuous function can always be constructed from a dataset. For example, we could construct a Baire space by distributing the observed categories on the standard basis on \(\mathbb{R}^{{C_n}-1}\). This construction is similar to the way that categorical variables are typically treated in statistical models, wherein each category is associated with a unique dimension in a high-dimensional space. Once we have created this mapping, we can define \(\beta(x) = \sum_{c=1}^{C_n-1} {\beta_{c} x_c}\). Here \(\{x_c\}_{c=1}^{C_n-1}\) is the representation of \(x\) on the standard basis on \(\mathbb{R}^{{C_n}-1}\), the \(C_n^{\text{th}}\) category is the reference category, and \(\beta_{c}\) is the coefficient on dimension \(c\).

Such a high-dimensional, orthogonal representation might not be the most efficient way to capture the relationships between categories. Specifically, in many types of qualitative data, the categories might be based on underlying low-dimensional properties. For example, colors are combinations of primary colors, human speech comprises basic phonetic sounds, and brands derive from elements of brand identity \citep{keller2020consumer}. In these examples, the minimal complete pseudometric space \(\mathcal{X}\) that satisfies our properties will likely be lower in rank than \(\mathbb{R}^{{C_n}-1}\). It is in such contexts and cases that our research is centered.

\subsection{Bounded Linear Bijective Operator}

We introduce an RKHS of representation functions, denoted by \( \mathcal{H} \). This space encapsulates the commonalities and variations among categories in the qualitative data, facilitating their seamless integration into the quantitative model. Furthermore, we define a bounded linear bijective operator, denoted by \( T: \mathcal{X} \rightarrow \mathcal{H} \). We introduce this operator to simplify the specification of a linear functional \( L \) on \( \mathcal{H} \):
\begin{equation}
L(h) = \beta(T^{-1}(h)), h \in \mathcal{H}. \label{eqn:f_T}
\end{equation}

The operator \( T \) acts a bridge, connecting the category space \( \mathcal{X} \) and the representation space \( \mathcal{H} \). Since each representation in \( \mathcal{H} \) corresponds to a unique category in the pre-image space of \( \mathcal{H} \), the functional \( L \) is implicitly defined through the continuous and linear function \( \beta \). This function accounts for the conditional mean shift that occurs when each category supersedes the reference category. 

The graph of the operator \( T \), denoted as \( G(T) = \{(x, T(x)): x \in \mathcal{X}\} \), depicts the Cartesian product of two distinct representations of qualitative data. The first is the abstract representation in \( \mathcal{X} \), which endows the observed discrete categories with a continuous representation. The second is the functional representation in \( \mathcal{H} \). The focal element in Equation \eqref{eqn:f_T} serves as an embedding of \( \mathcal{X} \), an extended set of categories equipped with a pseudometric determined by the outcomes, into \( \mathcal{H} \). Therefore, we refer to \( T \) as the Hilbert space embedding of the categories.

\subsubsection{Model Reformulation}

The properties of \( T \) and \( \beta \) imply that \( L \) is continuous and linear. Specifically, \( L = \beta \circ T \) is a composition of the continuous functions \( \beta \) and \( T \), thus it is also continuous. For all \( h_1, h_2 \in \mathcal{H} \) and all scalars \( c \), we have:
\begin{align*}
L(c h_1 + h_2) &= \beta(T(c h_1 + h_2)) \\
&= \beta(c T(h_1) + T(h_2)) & \text{(since $T$ is linear)} \\
&= c \beta(T(h_1)) + \beta(T(h_2)) & \text{(since $\beta$ is linear)} \\
&= c L(h_1) + L(h_2).
\end{align*}

Therefore, by the Riesz representation theorem (for a proof see \citealp{akhiezer2013theory}), there exists a unique vector \( p_L \in \mathcal{H} \) satisfying:
\begin{equation*}
L(h) = \langle h, p_L \rangle_H.
\end{equation*}
\noindent where $p_L$ is the Riesz representer of \( L \). 

This element of the representation space fulfills a role similar to $\beta$ in Equation \eqref{eqn:eqn2a} and enables us to refine Equation \eqref{eqn:eqn1a} as follows:
\begin{equation}
y_n = \alpha + \langle h_n, p_L \rangle_H + g(v_n; \zeta) + \epsilon_n, \label{eqn:eqn3a}
\end{equation}
\noindent where $h_n$ denotes the element in \( \mathcal{H} \) corresponding to observation $n$.

Our construction is informed by recent proposals for the use of Riesz representers in causal estimation \citep{bennett2022inference, chernozhukov2021automatic, chernozhukov2022riesznet}, yet it deviates significantly. Specifically, prevalent methodologies primarily focus on estimating a single linear functional of a high-dimensional function, such as the classic treatment effect model. In this context, the domain is a high-dimensional function that represents the difference in conditional expectations given treatment and control, and the codomain is the real line, symbolizing the treatment effect.

Unlike the cited papers that focus on single-valued functionals, our paper studies potentially infinite categories embodied in qualitative data. Consequently, while these methodologies apply to our model in the simplest case, where one category is represented in the qualitative data, they do not apply more generally.

As previously highlighted, the classical literature comprehensively addresses situations involving low-dimensional categories, but these are often ill-suited for many qualitative data applications. Therefore, we instead focus on scenarios where the qualitative data is complex, leading to the classical fixed effects estimator becoming either inadmissible or inherently inefficient to the point of impracticality. Moreover, even when the functional approach in extant papers can be extended to a vector-valued functional, the mathematical results would necessitate the codomain of the functional to be finite-dimensional, and likely even of fixed dimensionality. Our method uniquely accommodates the case of dynamically growing sets of categories through category emergence, a key consideration in our target applications.

\subsection{Constructing the Representation Map}

RKHSs possess several fundamental properties that have been extensively discussed in the literature \citep{ scholkopf2002learning}. One of the pivotal properties is that for an RKHS defined by the kernel \( k \),
\begin{equation*}
h(x) = \langle h, k(x, \cdot) \rangle,
\end{equation*}
\noindent for any point \( x \) in the domain and any function \( h \) within the RKHS \citep{aronszajn1950theory}. Building on this property, the value of \( p_L \)---the Riesz representer of a continuous linear functional \( L \)---at point \( x \) is expressed as:
\begin{equation*}
p_L(x) = \langle p_L, k(x, \cdot) \rangle.
\end{equation*}

Taking \( k(x, \cdot) \) as \( h \), we get:
\begin{equation*}
L(k(x, \cdot)) = \langle k(x, \cdot), p_L \rangle.
\end{equation*}

From this, we can conclude that \( p_L(x) = L(k(x, \cdot)) \). In the context of our model, the Riesz representer \( p_L \) of the functional \( L \) defined on \( \mathcal{H} \) evaluated at point \( x \) provides insights into how an outcome changes when the category associated with \( k(x, \cdot) \) replaces the reference category. If the Hilbert space embedding, denoted as \( T \), aligns with the feature map (i.e., \( x \to h \implies h = k(x, \cdot) \)), two significant implications emerge:
\begin{enumerate}
    \item \textbf{Consistency}: Every category in the Baire space maps back to itself through \( T^{-1} (k(x, \cdot)) \).
    \item \textbf{Practicality}: We can efficiently construct \( T \).
\end{enumerate}

Our focus is on models of this nature. To bridge the conceptual framework with its implementation, we introduce an initial injective mapping, \( \Gamma: \mathcal{X} \to \mathcal{Z} \), from the Baire space of categories \( \mathcal{X} \) to a numerical representation in a complete inner product space, \( \mathcal{Z} \). 

Examples of \( \Gamma \) can be found across a wide range of domains. In color theory, an RGB model can act as \( \Gamma \), mapping color categories into numerical representations. In audio processing, the Fourier transform can serve a similar purpose, mapping sound categories into a frequency representation. In more abstract scenarios, like the categorization of Netflix's micro-genres, the textual descriptions serve to define the categories. Phrases such as `critically-acclaimed emotional underdog movies' and `British set in Europe Sci-Fi \& Fantasy from the 1960s' can be processed through a sophisticated language model to generate numerical representations residing in $\mathcal{Z}$. Many traditional models and embedding algorithms map non-numerical data onto bounded convex subsets within the Euclidean space, using the dot product (often interpreted as the cosine distance). Such algorithms would be candidates for $\Gamma$.

Although initially designed for other applications, we repurpose these embeddings in our model using transfer learning. To refine these representations, we introduce a subsequent feature map, \( \phi: \mathcal{Z} \to \mathcal{H} \). In line with traditional kernel methods, we implicitly define 
\( \phi \) using the kernel function \( k \), thereby avoiding direct computation:
\begin{equation*}
\phi (x) = k(x, \cdot),
\end{equation*}
\noindent for any point \( x \) in the domain.

We focus on \( \mathcal{H} \) that can be constructed through the kernel trick, and where we can further refine Equation \eqref{eqn:eqn3a} to:
\begin{equation}
y_n = \alpha + \sum_i \alpha_i K (z_n, w_i) + g(v_n; \zeta) + \epsilon_n. \label{eqn:eqn4a}
\end{equation}
\noindent Here $z_n$ denotes the element in $\mathcal{Z}$ corresponding to observation $n$, while $\alpha_i, w_i$ denote weight vectors.

Equation \eqref{eqn:eqn4a} provides identifiable and robust estimates, even in scenarios where Equation \eqref{eqn:eqn1a} might be underidentified due to the unrestricted cardinality of the parameter vector \( \beta = {\beta_1, \dots, \beta_{C_n}} \). Specifically, $C_n$ could increase with $n$ at a rate that potentially disrupts inference. This is particularly likely in dynamic settings, such as longitudinal studies, where new categories continually emerge, and older categories become obsolete. Unlike \( \beta \), the weight vectors \( \alpha_i, w_i \) possesses a fixed cardinality that depends solely on the complexity of the categories, irrespective of $n$. We refer to $w_i$ as the weight vector to distinguish it from the structural parameters in the original model formulation, and because in the case where $K$ is the inner product in $\mathbb{R}^d$, $w_i$ acts as a set of weights on the dimensions of the category representations in $\mathcal{Z}$. In other cases, $K$ may have a distinct interpretation. For instance in the case of a radial basis function, $w_i$ may be the center point, yielding the `ideal point' model \citep{srinivasan1973linear} and the multiple `ideal point' model \citep{lee2002multiple}.

It is important to underscore that our proposed model framework is tailored for causal inference. The primary construct of interest is the impact of qualitative variables, represented through the Hilbert space, on the outcome variable. In line with prior work, solutions to risk minimization problems involving both an empirical risk and a regularizer are often expressible as expansions in terms of the training examples \citep{kimeldorf1971some, scholkopf2001generalized}. We can effectively apply this principle to Equation \eqref{eqn:eqn4a} by aligning the loss function with the error term as indicated in Equation \eqref{eqn:eqn1b}, and integrating an appropriate regularizer like the quadratic regularizer. In these circumstances, an optimal solution resides within the linear hull of the training examples as mapped into the feature space. An approximation to this solution can be computed effectively utilizing well-established methodologies and solutions, circumventing the need to explicitly compute the fixed effects.

\subsection{Nonparametric Model}
\label{section:nonparametric}
It is feasible to construct $\mathcal{X}$ and $\beta$ within a nonparametric model of qualitative information. Specifically, suppose Equation \eqref{eqn:eqn1a} takes the form:

\begin{equation*}
y_n (I_n, v_n) = \alpha + h (I_n, v_n; \eta) + \epsilon_n,
\end{equation*}

\noindent where \(h\) is Lipshitz continuous and smooth, and \( \eta \) incorporates the effects captured by \( \beta \) and \( \zeta \) in the semiparametric form. We can arrange the observed categories on the standard basis, and $v_n$ on additional dimensions, such that $\beta \equiv h$. This construction is isometric; there's no distortion in the embedding of categories and independent variables, even in the case of a fairly general $h$ where we may otherwise have anticipated distortion, because we have access to $\beta$ in addition to \(\mathcal{X}\), and the pseudometric on $\mathcal{X}$ is defined by applying the Euclidean metric to $\beta(x)$ for $x \in \mathcal{X}$. It is for these features that we specify $\mathcal{X}$ as a Baire space, as it enables the distortion-free embedding of $C$ categories and any additional independent variables in the `flat' space of $\mathcal{X}$ for a wide range of functional forms.

However, even though the nonparametric approach is mathematically appealing, it may be challenging to estimate it in the kinds of data for which our model is designed. Specifically, in typical applications, we expect data sparsity and categories to become outdated. Under these conditions, we hypothesize that the parsimonious linear and additive form of the traditional semi-parametric linear model may outperform more flexible specifications in a majority of applications.

\subsection{Model Estimation Procedure}

Previous sections established this model's theoretical consistency with qualitative data, a known challenge for conventional econometrics, and its adaptability through diverse kernel functions that represent categorical data. This section details practical implementation, highlighting the integration of transfer learning and established econometric techniques like regression.

\begin{enumerate}

\item \textbf{Baseline Representation:} Choose a baseline representation using a mapping \( \Gamma \) and kernel \( K \) within the RKHS framework. \( \Gamma \) could leverage existing models, transfer learning, or feature extraction. The model is a linear regression with \( \alpha \) representing the baseline category and \( \sum_i \alpha_i K (z_n, w_i) \) a linear additive term.

\item \textbf{Determination of \( \alpha_i \) and \( w_i \):} Estimate \( \alpha_i \) and \( w_i \) based on \( z_n \), \( \Gamma \), and \( K \). Potential estimation techniques include Maximum Likelihood Estimation, Bayesian methodologies, or the Method of Moments. For linear kernels, \( z_n \) can be treated as covariates in the regression, with \( \sum_i {\alpha_i w_i} \) as coefficients.

\item \textbf{Kernel Function Selection:} When applying the kernel trick, it is important that the kernel \(K\) satisfies Mercer's theorem: \(K\) must be positive definite. This property ensures that \(K\) corresponds to an inner product in some feature space induced by a feature map \(\phi\).

\item \textbf{Fixed Effects Calculation:} For category \( c \), calculate the fixed effect, denoted \( \beta(c) \), using the equation \( \beta(c) = \sum_i \alpha_i K (z(c), \hat{w}_i) \), where \( z(c) \) represents category \( c \) in \( \mathcal{Z} \). \( \hat{w}_i) \) are the inferred weights.

\end{enumerate}

\section{Empirical Process}
\label{section:empirical_process}

In this section, we introduce and illustrate a novel data generating process that uncovers key limitations of traditional fixed effects regressors in handling complex categorical data. Through an analogy we term `Fido's Ball', we employ the Yule-Simon process to highlight scenarios where conventional approaches struggle, underscoring the need for our proposed model that handles such complex categorical data more effectively. We delve into the empirical process to expose the underlying issues impacting traditional techniques, including data sparsity and the emergence of new categories. Subsequent sections will empirically demonstrate these challenges through simulations and a real-world dataset, further revealing the advantages of our method, as well as establishing the convergence and stability of our proposed model and estimator.

\subsection{Yule-Simon Process}

The Yule-Simon model is a statistical process that yields a distribution that is particularly suited to representing natural phenomena with many rare events. We consider the following setup, as described and analysed by \cite{bertoin2020linear}. To aid comprehension and facilitate further exploration, we intentionally adopt the same notation as Bertoin. This provides an opportunity for interested readers to delve deeper into the theoretical results in Bertoin's paper as a means of expanding upon our own findings.

Consider $U_1, U_2, \ldots$ as i.i.d. uniform random variables on the interval $[0, 1]$. We define the sequence of (uniform) empirical processes as:
\begin{equation*}
  G_n(u)=\frac{1}{\sqrt{n}}\sum_{i=1}^{n}\left(1_{U_i\leq u}-u\right),\ u \in[0,1],
\end{equation*}
\noindent where this sequence converges in distribution as $n\rightarrow\infty$ towards a Brownian bridge $(G(u))_{0\leq u\leq 1}$ in the Skorokhod metric.

To illustrate the empirical process, consider a sequence $\varepsilon_2,\varepsilon_3,\ldots$ of i.i.d. Bernoulli variables with a fixed parameter $p\in(0,1)$ that signify when repetitions occur, such that the $n^{\text{th}}$ step of the algorithm is a repetition if $\varepsilon_n=1$, and an innovation if $\varepsilon_n=0$. For every $n\geq2$, we also define $v(n)$ as a uniform random variable on the set $\{1,\ldots,n-1\}$, independent of the other variables, which specifies which preceding item is copied when a repetition happens.

We set $\varepsilon_1=0$ for definiteness and recursively construct a sequence of random variables $\hat{U}_1,\hat{U}_2,\ldots$:
\begin{equation*}
  \hat{U}_n=
  \begin{cases}
    \hat{U}_{v(n)}, & \text{if } \varepsilon_n=1, \\
    U_{i(n)},       & \text{if } \varepsilon_n=0,
  \end{cases}
\end{equation*}
\noindent where
\begin{equation*}
  i(n)=\sum_{j=1}^{n}(1-\varepsilon_j)\quad\text{for } n\geq1
\end{equation*}
\noindent denotes the total number of innovations after $n$ steps. It is implicit that the sequences $(v(n))_{n\geq2}$, $(\varepsilon_n)_{n\geq2}$, and $(U_j)_{j\geq1}$ are independent.

This represents a linear reinforcement procedure (i.e., a preferential attachment process), implying that, provided $i(n)\geq j$ (i.e., the variable $U_j$ has already appeared at the $n^{\text{th}}$ step of the algorithm), the probability that $U_j$ is repeated at the $(n+1)^\text{th}$ step is proportional to the number of its previous occurrences. We henceforth refer to the parameter p of the Bernoulli variables $\varepsilon_n$ as the reinforcement parameter. While each variable $\hat{U}_n$ adheres to the uniform distribution on $[0, 1]$, the reinforced sequence $(\hat{U}_n)_{n\geq1}$ is not stationary, exchangeable, or even partially exchangeable. In what follows, we refer to realizations of $(\hat{U}_n)_{n\geq1}$ as Yule-Simon draws.

Moreover, while the conclusions of the Glivenko–Cantelli theorem apply, the conclusions of the Donsker theorem do not \citep{bertoin2020linear}. Specifically, for certain parameter values, the sequence of empirical processes $(\hat{G}_n(u))_{0\leq u\leq 1}$ defined in Equation \eqref{eqn:IndF} converges in law to a Brownian bridge up to a scale factor, whereas for other parameter values, an additional rescaling is necessary and the limit is a Brownian bridge with exchangeable increments and discontinuous paths.

\subsection{Fido's Ball}
\label{sec:fido}

We describe our proposed empirical process through the following analogy. Consider an ongoing game of fetch between Fido, a `memoryless' Golden Retriever, and his owner, Odif. This serves as an illustration of the empirical process. In each iteration of this never-ending game, Odif selects a ball using the Yule-Simon process. Though the balls are inherently unordered, they are organized using an injective map from the interval $[0, 1]$ to the type (i.e. category) of ball. This arrangement allows a Yule-Simon draw to specify ball type. Importantly, the process does not require the selection of distinct ball types; many, or potentially all, values within the closed interval may map to the same type. Nonetheless, our research is most relevant to scenarios where the codomain of the map is a diverse and potentially infinite set of ball types.

During each iteration, Fido chases and retrieves the ball thrown by Odif. The joy Fido experiences in the $i^{th}$ round depends on the type of ball thrown:
\begin{equation}
\mathrm{joy}_i = \alpha + \beta (\mathrm{type}(\hat{u}_i)) + \epsilon_i, \label{eqn:Fido1}
\end{equation}
\noindent Here, $\beta$ is a mean-shift function. $\hat{u}_i$ is the $i^{\text{th}}$ Yule-Simon draw. $\mathrm{type}$ is a map from the closed interval $[0, 1]$ to ball types, such that $\mathrm{type}(u_i)$ denotes the type of ball thrown in iteration $i$.

We consider compositional maps of the form $\tau(\mathrm{type}(\hat{u}_i))$ from the closed interval $[0, 1]$ to some bounded and closed subset of $\mathbb{R}^d$, where $d$ is finite, that is continuous and linear. Specifically, in the context of color, consider the RGB color model with the slight modification that the intensity of the three primary colors is expressed in the unit interval and $\tau(\mathrm{type}): [0,1] \to [0,1]^3$. We use this map to define $\beta$ as follows:
\begin{equation*}
\beta (\mathrm{type}(\hat{u}_i)) = w_r \mathrm{red}(\hat{u}_i) + w_g \mathrm{green}(\hat{u}_i) + w_b \mathrm{blue}(\hat{u}_i),
\end{equation*}
\noindent where ${w_r, w_g, w_b}$ is a weight vector, and $\{\mathrm{red}, \mathrm{green}, \mathrm{blue}\}$ are constituent submaps of $\tau(\mathrm{type})$ such that each draw is matched to a category (i.e., color), which is then mapped to a vector of primary color intensities. Note that, although we use the linear kernel to construct $\beta (\mathrm{type}(\hat{u}_i))$ for the sake of clarity in this exposition, more complex kernels could also be employed.

As we discuss in Section \ref{sec:AA}, traditional models are likely to struggle to estimate Fido's joy empirically. First, an infinite number of categories can be introduced through $\mathrm{type}$ and the Yule-Simon process. This would require an infinite number of fixed effects, which would lead to concerns about the complexity of the parameter space. These concerns are likely to be amplified in parameter ranges where the rate of emergence of new categories is high as the rate of emergence of new categories may be faster than the rate at which information accumulates on the parameters associated with each category.

Crucially, the traditional estimator may still be consistent if the categories emerge rarely (i.e., the reinforcement parameter is close to 1). In such cases, it is likely that the categories that emerge late in the data may be reinforced rarely as the model places more weight on random variables that occured more often in the past. Therefore, in such cases, even if the new categories emerge slowly, the rate of emergence may be too great for the rare categories in the data where novel information is increasingly unlikely to accumulate with growth in the dataset.

In this paper, we pose the question: given that Odif is unaware of $\beta$, but cognizant of $\tau$ and $\mathrm{type}$ (with the former being a natural fact and the latter an ordering), can Odif use $\tau(\mathrm{type})$ to estimate Fido's joy? More generally, can the qualitative data inference problem be solved by deriving a map of the form of $\tau(\mathrm{type})$ to recast the qualitative data as a fixed-length vector of numerical representations that can then be incorporated in the statistical model?

We answer this question affirmatively. In the context of the linear kernel, we reframe Equation \eqref{eqn:Fido1} as:
\begin{equation}
\mathrm{joy}_i = \alpha + w_r \mathrm{red}(\hat{u}_i) + w_g \mathrm{green}(\hat{u}_i) + w_b \mathrm{blue}(\hat{u}_i) + \epsilon_i. \label{eqn:Fido3}
\end{equation}

Equation \eqref{eqn:Fido3} is straightforward to estimate and identify from the data. Notably, its analysis uncovers an essential identification condition in the linear model---the coefficients are only identified based on the variations induced in the data along the dimensions of the codomain of \(\mathrm{type}\). For instance, if Odif selects balls varying only in the primary color `red,' then the coefficients for `blue' and `green' become underidentified. This requirement is well established in the literature. A coefficient such as \(w_r\) can only be determined if variation exists in the corresponding dimension of the independent variables, e.g., the color of the ball. Moreover, this condition is naturally likely to be satisfied in relevant data, as the categories emerge uniformly in it, and are therefore equally likely to vary in each dimension of \(\mathrm{type}\) (i.e., `red' balls are equally likely as `green' or `blue' balls).

In the case of a non-linear kernel, the estimation is expressed as:
\begin{equation*}
\mathrm{joy}_i = \alpha + K ( \{\mathrm{red}(\hat{u}_i), \mathrm{green}(\hat{u}_i), \mathrm{blue}(\hat{u}_i) \}; w) + \epsilon_i.
\end{equation*}
Here \( w = \{w_r, w_g, w_b, \dots\} \) corresponds to the single component model, and we focus on models of this specific form. More intricate forms that seek a blend of kernels or adopt a compositional strategy in constructing advanced kernels can also be estimated through non-linear least squares (in the case of the specification of an empirical loss function) or likelihood-based methodologies (when delineating the distribution of the error term, among other model elements) provided the data has sufficient resolution to identify the more complex form.

\section{Model Validation Through Simulations}

Simulations play a crucial role in validating our proposed model, particularly when contrasting it with traditional approaches for estimating complex qualitative data. Our validation pursues two main objectives: (1) elucidating the challenges faced by traditional models in handling evolving categorizations and sparse data categories, and (2) revealing the robustness of our model, even in the face of swiftly proliferating new categories. We undertake these evaluations by generating data through the Yule-Simon process, characterized by the following:

\begin{enumerate}
\item Unique Yule-Simon draws form distinct categories.
\item The reinforcement parameter, ranging between 0.01 and 0.99, signifies the probability that an extant category recurs.
\item The value of each draw is the fixed effect of its category; hence, all fixed effects fall within the range of 0 to 1.
\item A continuous covariate follows an i.i.d. standard Normal distribution.
\item The error term also follows an i.i.d. standard Normal distribution.
\end{enumerate}

Subsequent subsections first detail the convergence dynamics for the conservative case, where the reinforcement parameter is set at 0.99. This setting implies that a new category emerges once every 100 observations, with the remaining 99 relating to existing categories. We explore a variety of reinforcement parameters, ranging from a very high rate of category emergence to the lower rate described in the previous simulation. Next, we illustrate the implications of estimation error on inference by showcasing parameter estimates. We compare the performance of variable selection methods, such as LASSO, and regularized parameter estimation methods like Ridge Regression, to traditional estimators on the simulated data. Finally, we present our proposed model and conclude.

As a brief preview, we demonstrate that traditional models struggle to provide consistent estimates when dealing with complex, evolving qualitative data. In contrast, our proposed model maintains robust performance even when new categories emerge rapidly, validating its ability to effectively handle data-sparse categories and shifting categorization systems. The method's superior consistency, precision, and computational efficiency make it a modern alternative to traditional models for qualitative data analysis.

\subsection{Convergence Dynamics}

In this subsection, we explore the convergence dynamics of the traditional estimator by generating simulated datasets of varying sizes, ranging from \(10^{2.5}\) to \(10^5\), spaced evenly on a logarithmic scale. This broad range provides a comprehensive view of the estimator's performance across different scales, capturing both nuances and overarching trends. Our analysis focuses on two key metrics: estimation error and entropy. The absolute estimation error measures the accuracy in recovering true fixed effects, while entropy gauges uncertainty in the inferred estimates. Together, these metrics illuminate the traditional estimator's efficacy in assimilating information.

\subsubsection{Estimation Error}

\begin{figure}[tbp]
\centering
\includegraphics[width=0.75\linewidth]{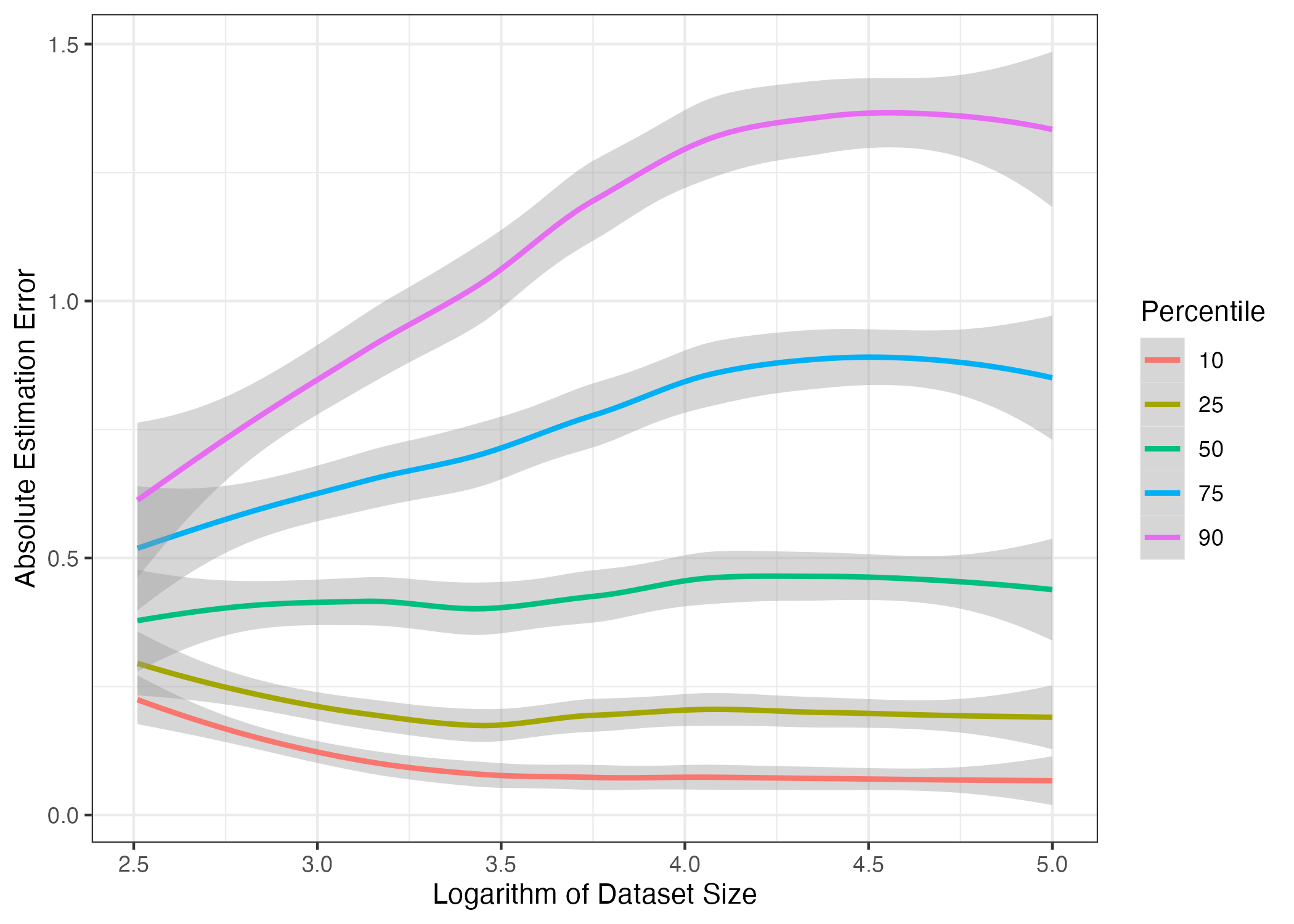}
\caption{Absolute Estimation Error and Dataset Size}
\begin{minipage}{\linewidth}
\medskip
\footnotesize
Note: The reinforcement parameter is set at 0.99. The dataset size is represented in base 10 logarithm.
\end{minipage}
\label{fig1:estimates_size}
\end{figure}

Under classical conditions, the discrepancy between true and estimated values should decrease as dataset size increases, reflecting the assimilation of information. To investigate this, we computed the absolute estimation error for each category in the simulated data as the absolute difference between the estimated and actual values. We identified the 10\textsuperscript{th}, 25\textsuperscript{th}, 50\textsuperscript{th}, 75\textsuperscript{th}, and 90\textsuperscript{th} percentiles of these errors and plotted them against the logarithm (base 10) of the dataset size in Figure \ref{fig1:estimates_size}.

Figure \ref{fig1:estimates_size} reveals two principal observations. Initially, in smaller datasets, estimation errors remain consistent across percentiles, a predictable outcome given the restricted range of categories and limited observations. However, as the dataset grows, a clear divergence occurs. Errors for the lower percentiles decrease, reflecting the incorporation of new information in the more commonly observed categories, while errors for the higher percentiles increase due to the emergence of new, infrequently reinforced categories, leading to imprecision. Strikingly, the median absolute error remains relatively stable at around 0.5 across dataset sizes. Considering that true fixed effects range between 0 and 1, this consistent median error underscores a potentially non-informative estimation process.

The widening gap in errors between frequently and rarely observed categories highlights the challenges that conventional regression techniques face when handling intricate categorical data. Categories that appear later in the data, particularly those that are seldom observed, do not show improved estimates as the dataset grows. This inefficiency, especially when faced with swiftly emerging and underrepresented categories, underscores a critical challenge and emphasizes the need for innovative approaches that are adaptive to the dynamic nature of category emergence and representation.

\subsubsection{Entropy}

We proceed to calculate the entropy of the inferred estimates derived from the traditional regression estimator. The entropy of a $D$-dimensional multivariate Gaussian with covariance matrix $\Sigma$ is given by:
\begin{equation}
H(x_{MVN}) = \frac{D}{2} \left( 1 + \log (2 \pi) \right) + \frac {1}{2} \log |\Sigma|. \label{eqn:entropy}
\end{equation}

Setting aside the Donsker conditions, the inferred distribution of parameters converges to a multivariate Normal with estimable asymptotic mean and covariance using standard techniques. Furthermore, in our simulations, the regression estimator is efficient since the error term follows a standard Normal distribution. Thus, we compute the entropy of the inferred parameter distribution based on this functional relationship.

\begin{figure}[tbp]
\centering
\includegraphics[width=0.75\linewidth]{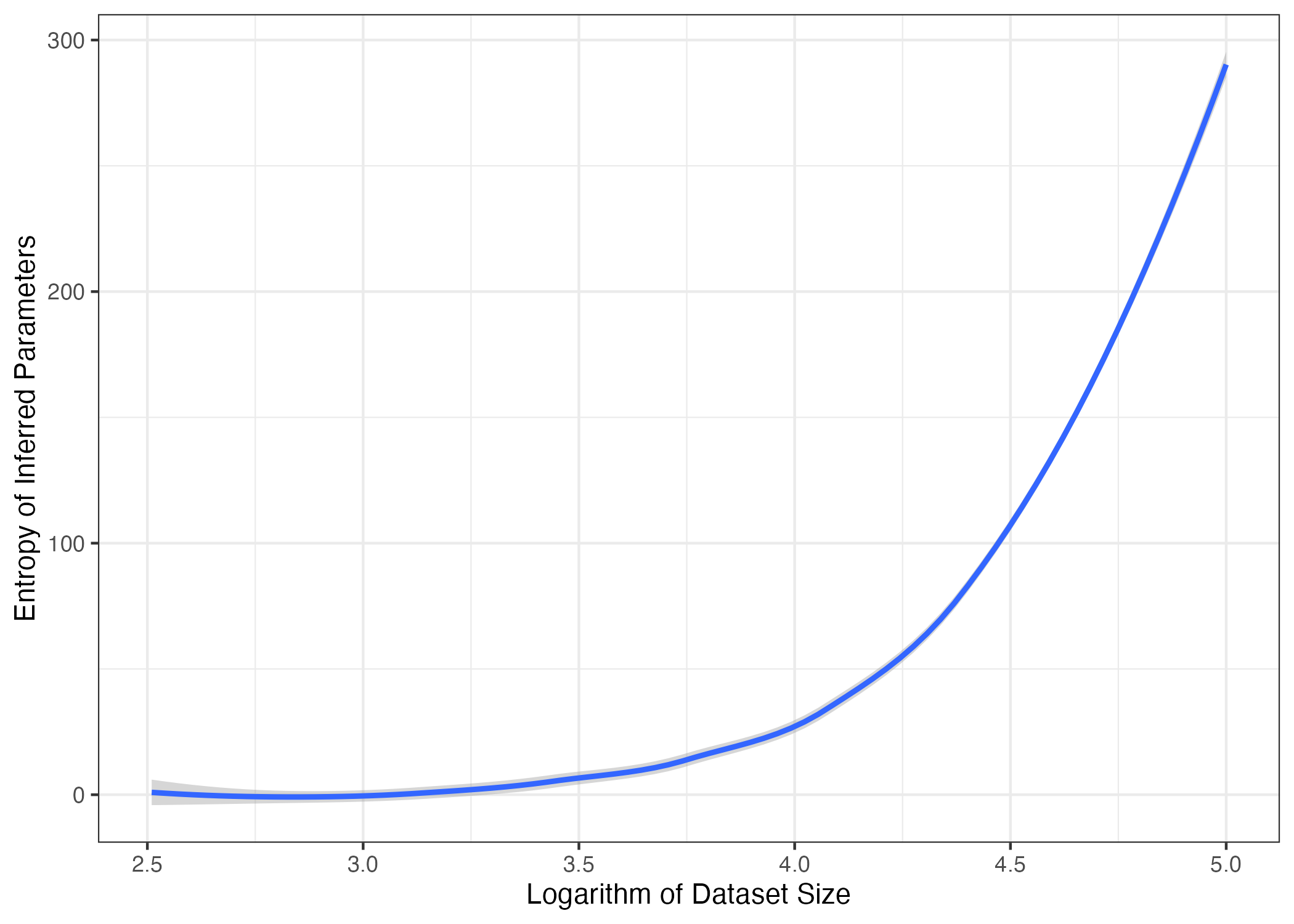}
\caption{Entropy and Dataset Size}
\begin{minipage}{\linewidth}
\medskip
\footnotesize
Note: The reinforcement parameter is set at 0.99; the dataset size is represented in base 10 logarithm.
\end{minipage}
\label{fig2:entropy_size}
\end{figure}

Figure \ref{fig2:entropy_size} shows that the entropy of the inferred parameters increases with the sample size, a trend driven by the introduction of new but infrequently reinforced categories within the data. In a scenario where the reinforcement process were to favor newly emerging categories, the model might preserve consistency through the aggregation of information. However, our simulations reveal a different behavior. Specifically, the increase in entropy from the first term in Equation \eqref{eqn:entropy} is not adequately offset by the decrease in the second term, causing a divergence in the values. This divergence not only highlights the limitations of conventional modeling techniques when dealing with rapidly evolving categories but also emphasizes the necessity for more sophisticated methods that can accurately interpret and represent the inherent complexity of such categorical data.

\subsection{Reinforcement Parameter}
\label{sec:reinforce_par}

Our initial exploration utilized a reinforcement parameter of 0.99, signifying a context where new categories arise infrequently. To deepen our understanding, we extended our study to investigate a spectrum of reinforcement parameter values, holding the dataset size constant at 5,000 observations to ensure a clear manifestation of the data's long-term patterns.

We systematically generated data for reinforcement parameters, incrementing by 0.01, spanning from 0.01 to 0.99. A parameter of 0.01 implies a high frequency of new categories, whereas 0.99, as previously studied, establishes a conservative testing ground for the regression model.

\begin{figure}[tbp]
\centering
\includegraphics[width=0.75\linewidth]{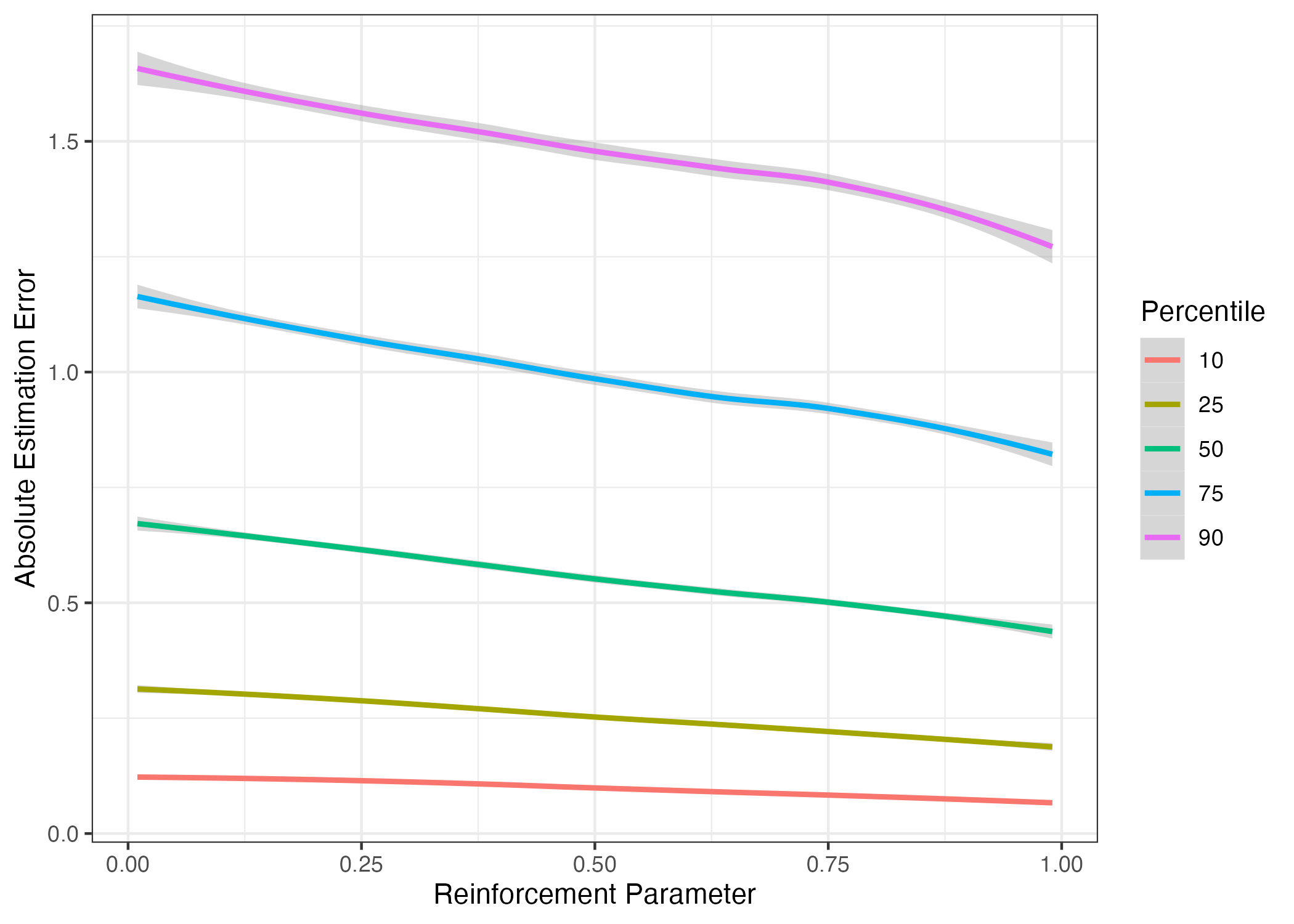}
\caption{Absolute Estimation Error and Reinforcement Parameter Values}
\begin{minipage}{\linewidth}
\medskip
\footnotesize
Note: Dataset size is 5,000; local polynomial regression fits for various error percentiles are shown in different colors; standard error bands are represented in grey.
\end{minipage}
\label{fig3:parameters}
\end{figure}

Figure \ref{fig3:parameters} shows that estimation errors are notably higher for lower reinforcement parameters. Surprisingly, even at higher values, where the resulting data is similar in category frequency to many observed datasets and where one might expect less error, significant deviations persist. This divergence from the expected root-n convergence can be traced to two fundamental processes: (1) the emergence of new categories, where lower reinforcement values usher in a surge of new categories, saturating the model; and (2) the reinforcement dynamics of existing categories, where the preferential attachment process makes gaining reinforcement for sporadically reinforced categories challenging, thereby inducing data sparsity.

\subsection{Parameter Estimates}

\begin{figure}[tbp]
\centering
\includegraphics[width=0.75\linewidth]{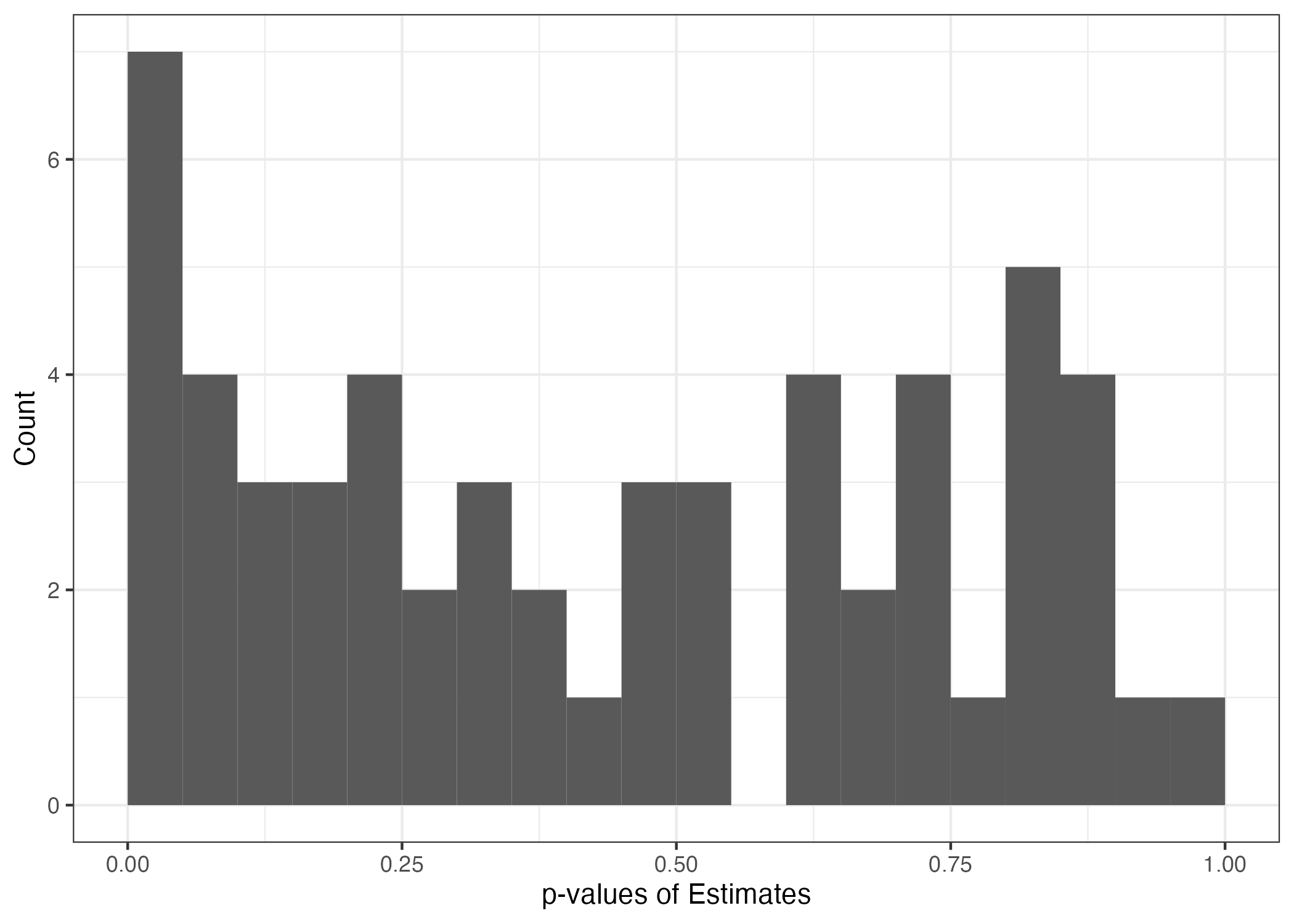}
\caption{Histogram of Regression P-values}
\begin{minipage}{\linewidth}
\medskip
\footnotesize
Note: Dataset size is 5,000; reinforcement parameter is 0.99. $p$-values of estimates from a standard regression.
\end{minipage}
\label{fig4:pvalues}
\end{figure}

Expanding on our prior examination of the reinforcement parameter's impact on estimation accuracy, we now delve into the credibility of our derived estimates. Figure \ref{fig4:pvalues} showcases the p-values of estimates for the same simulated dataset as in Section \ref{sec:reinforce_par} that has 5,000 observations and a reinforcement parameter of 0.99. Despite all fixed effects in our simulation being non-zero, the results are alarming: 87.72\% of the cases yield a p-value greater than 0.05, and 80.7\% surpass 0.1. This indicates that in many cases, the estimated effects are statistically no different from the null hypothesis, casting doubt on the reliability of traditional estimation methods in complex, real-world contexts.

Our dataset comprises a total of 58 parameters: 57 fixed effects and one continuous covariate. Accepted guidelines, like the rule of thumb suggesting at least 10 data points for each estimated parameter \citep{harrell1984regression}, as well as more stringent benchmarks such as 1 in 20 or 1 in 50 observations per variable \citep{steyerberg2000prognostic}, would deem our sample size to be sufficiently large. However, instead we find that the data is insufficient to obtain reliable estimates for the vast majority of parameters using the traditional estimator. This discrepancy underscores the limitations of the classical model and cautions against undue reliance on conventional benchmarks.

\begin{figure}[tbp]
\centering
\includegraphics[width=0.75\linewidth]{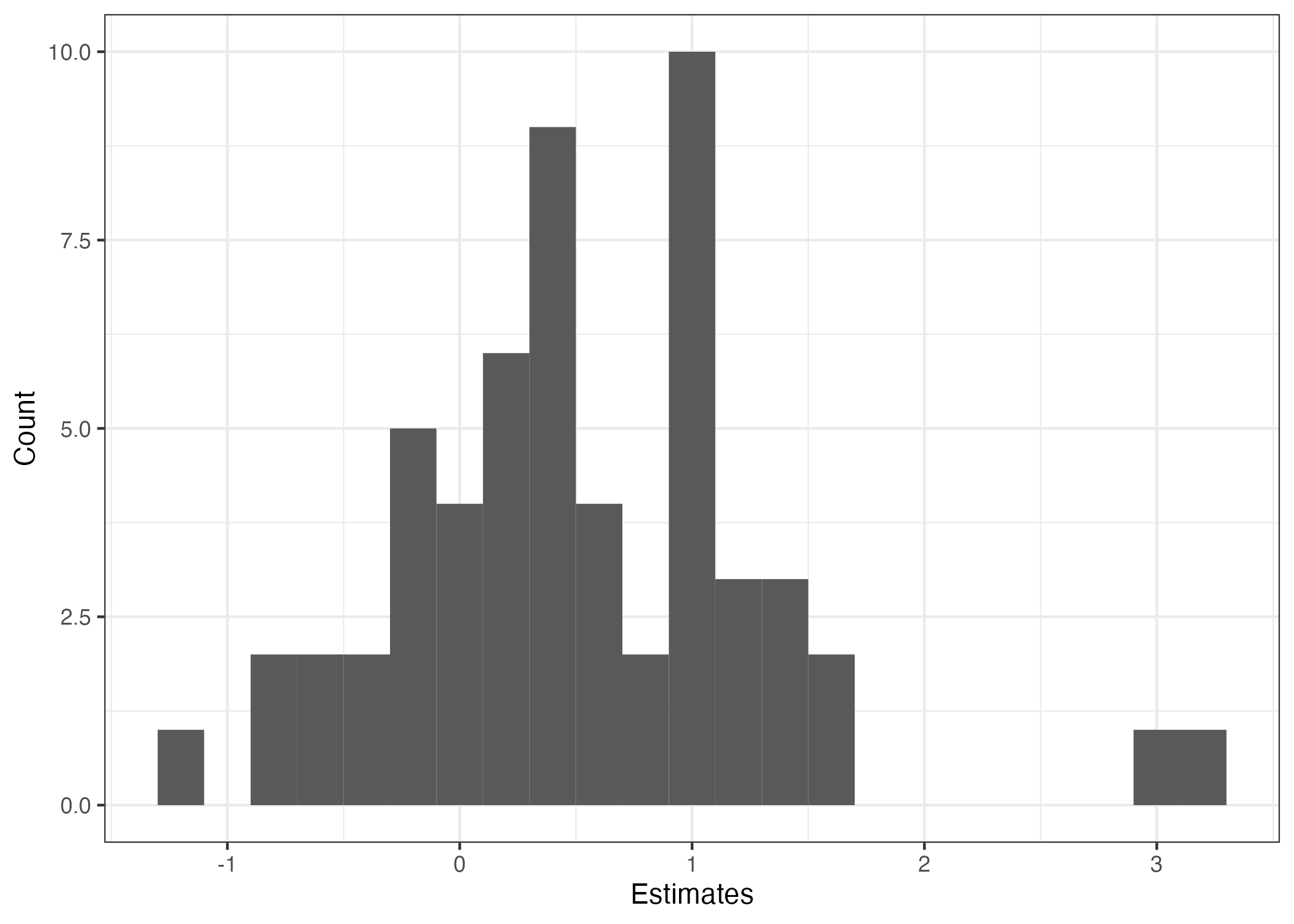}
\caption{Histogram of Regression Estimates}
\begin{minipage}{\linewidth}
\medskip
\footnotesize
Note: Dataset size is 5,000; reinforcement parameter is 0.99. Estimates from a standard regression.
\end{minipage}
\label{fig5:estimates}
\end{figure}

Delving into the estimates, Figures \ref{fig5:estimates} depicts a histogram of the estimates. It is concerning to observe that nearly half (27 out of 57, or 47.4\%) of these estimates fall outside their true value range, which is between 0 and 1. Such overextension aligns with our earlier concerns regarding data sparsity. Moreover, examining the significant estimates reveals a tendency for them to be large in magnitude, with 5 out of 7 falling outside the [0, 1] range and 1 falling almost at the boundary (0.987). This occurs because smaller effect sizes require more statistical power to resolve. Therefore, it is the larger estimates that are more likely to be significant, either because the underlying effect size is large or due to noise. These findings highlight how models can be misled by sparse categorical data, raising serious questions about the validity of existing estimation techniques. The results underscore the urgency of reevaluating conventional benchmarks and exploring alternative methodologies, as detailed in the subsequent sections.

\subsection{LASSO}

LASSO, a prominent variable selection method, provides promising solutions for confronting the estimation challenges and overfitting risks that are prevalent in traditional estimators for high-dimensional categorical data. Leveraging regularization procedures, LASSO distinguishes between influential and non-essential predictors, focusing on the selection of the most pertinent variables. By systematically penalizing the magnitudes of coefficients and nullifying those of less impactful variables, LASSO creates a more concise and interpretable model. When applied to dummy variable representations of categories, LASSO presents the potential to differentiate the truly impactful categories from the non-essential ones, effectively addressing the limitations observed in conventional regression methods within our simulated dataset.

\begin{figure}[tbp]
\centering
\includegraphics[width=0.75\linewidth]{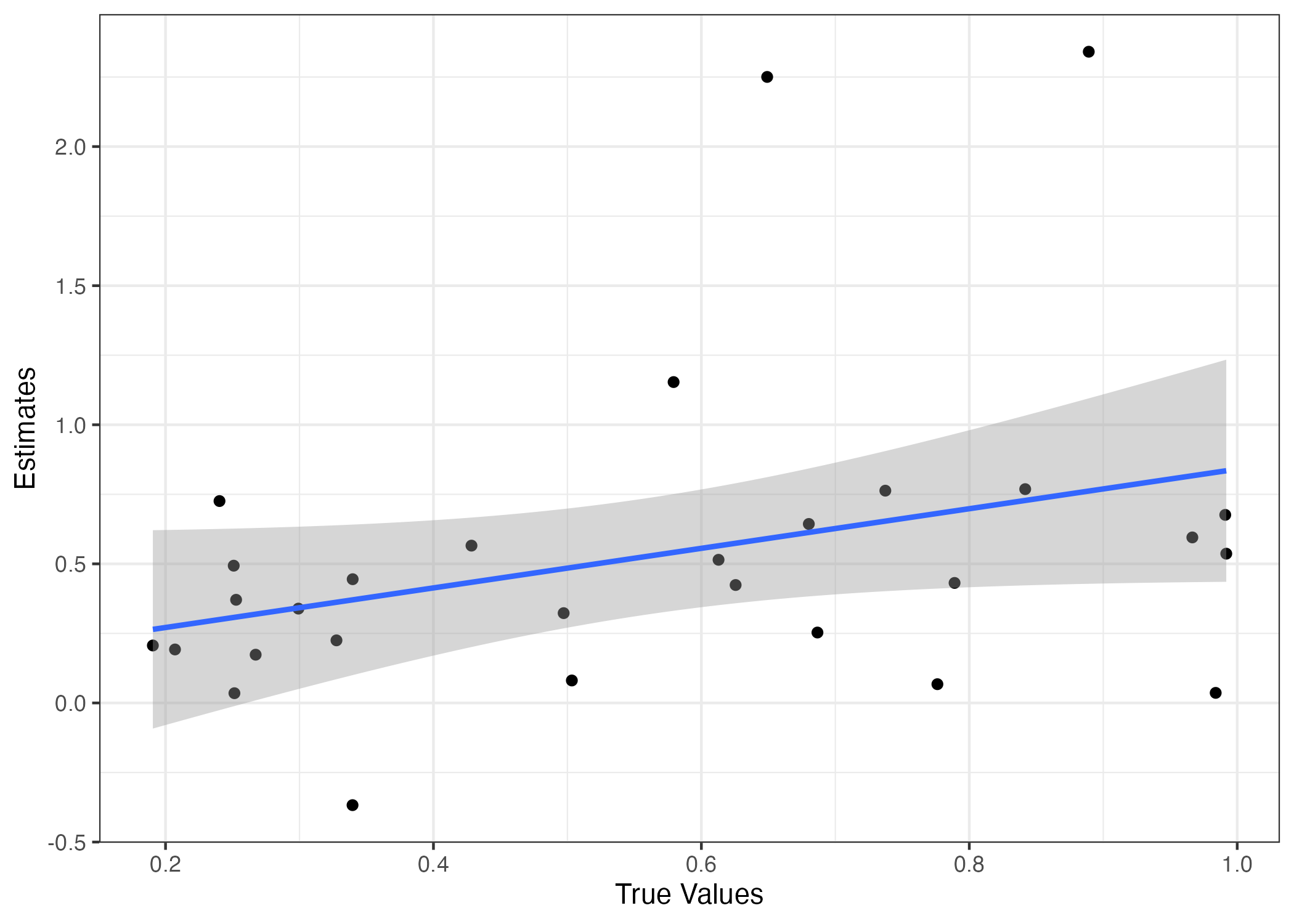}
\caption{Scatter Plot of LASSO Estimates}
\begin{minipage}{\linewidth}
\medskip
\footnotesize
Note: Dataset size is 5,000; reinforcement parameter is 0.99. Regression fit in blue; standard error bands in grey.
\end{minipage}
\label{fig7:LASSO_estimates}
\end{figure}

From the 57 fixed effects, LASSO provides estimates for only 29 in our 5,000-observation dataset, signifying an average drop of about half the fixed effects. Each retained effect is backed by 170 observations. Despite this seemingly substantial support, the derived estimates differ significantly from the actual values. Figure \ref{fig7:LASSO_estimates} depicts our derived estimates against the true values; the significant deviation from the ideal straight line with a slope of 1 illustrates the disparity.

This analysis underscores LASSO's limitations in sparse categorical data contexts such as dynamic category structures and data sparsity. The method might falter in offering consistently accurate insights for retained categories and omit many entirely. This limitation arises from LASSO's reliance on observed data frequencies to determine variable importance, failing to discern latent relationships between categories that could promote information sharing.

\subsection{Ridge Regression}

Alongside LASSO, we explore Ridge Regression as an alternative regularization approach designed to address the complexities of high-dimensional models. Both LASSO and Ridge Regression aim to curtail overfitting and mitigate the estimation challenges inherent in high-dimensional data, with their main point of divergence residing in their penalization strategies.

LASSO, utilizing an L1 penalty, can lead to certain coefficients being completely nullified. This effect is particularly concerning in scenarios with sparse categorical data, as it may result in the omission of predictors with latent importance that is not readily apparent from mere data frequencies. In contrast, Ridge Regression adopts an L2 penalty, ensuring that all predictors remain in the model, albeit with potentially shrunken coefficients, and are never entirely eliminated. This differentiation is critical, especially in sparse categorical data contexts where it is known that the categories are likely to exert a non-zero influence, making Ridge Regression an appealing alternative for capturing nuanced category relationships often missed by LASSO.

The L2 penalty in Ridge Regression suggests that the optimal penalty parameter, \( \lambda \), might differ from its counterpart in LASSO. Therefore, as with LASSO, we employ cross-validation to pinpoint \( \lambda \).

\begin{figure}[tbp]
\centering
\includegraphics[width=0.75\linewidth]{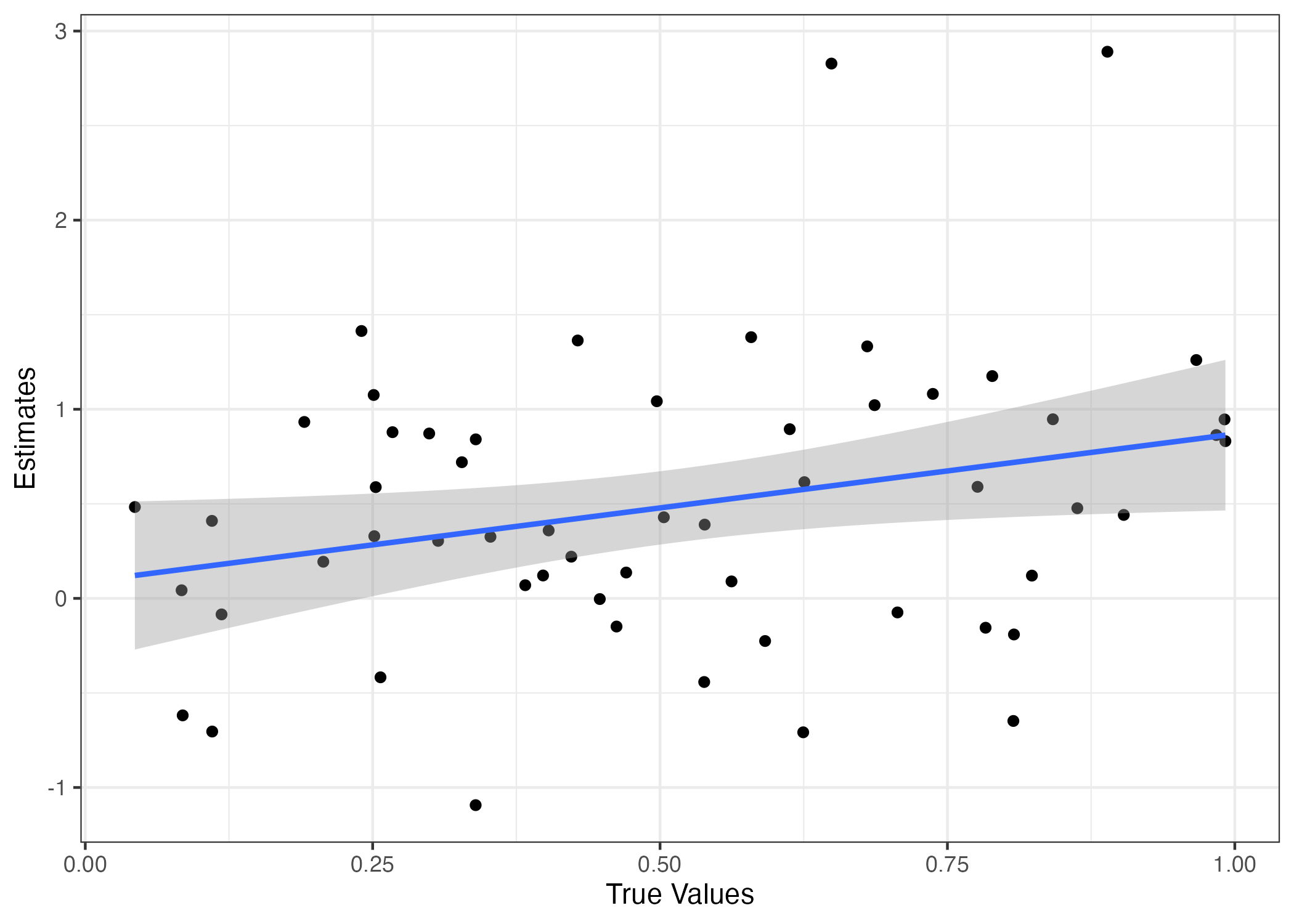}
\caption{Scatter Plot of Ridge Regression Estimates}
\begin{minipage}{\linewidth}
\medskip
\footnotesize
Note: Dataset size is 5,000; reinforcement parameter is 0.99. Regression fit in blue; standard error bands in grey.
\end{minipage}
\label{fig8:Ridge_estimates}
\end{figure}

The scatter plot in Figure \ref{fig8:Ridge_estimates} illustrates the relationship between our derived estimates from Ridge Regression and the actual values. While there's a discernible departure from an ideal line with a slope of 1, Ridge Regression offers an alternative perspective when juxtaposed with LASSO. This suggests that under specific conditions, Ridge Regression might be more adept at unveiling subtle relationships within the data.

Interestingly, the line illustrating the relationship between the estimated coefficients and the true values in both LASSO and Ridge Regression has a slope that closely approaches 0, with a positive intercept. This behavior indicates that both estimators, in an effort to minimize estimation variance, gravitate towards assigning the mean of the actual fixed effect values (0.5) to all fixed effects. Consequently, this strategy may render the estimates somewhat uninsightful, as the deviations of the category coefficients (i.e., the fixed effects) from the mean become relatively inconsequential.

\subsection{Rule-of-Thumb Aggregation}

To address the complexity posed by a vast number of categories, another viable strategy is category aggregation. This method entails combining several low-frequency categories into a broader, unified category. As a general guideline, we merge any category containing fewer than 5 observations, given the low probability of accurate estimation for such categories. Similar rule-of-thumb-based aggregation strategies are prevalent in the social sciences literature.

\begin{figure}[tbp]
\centering
\includegraphics[width=0.75\linewidth]{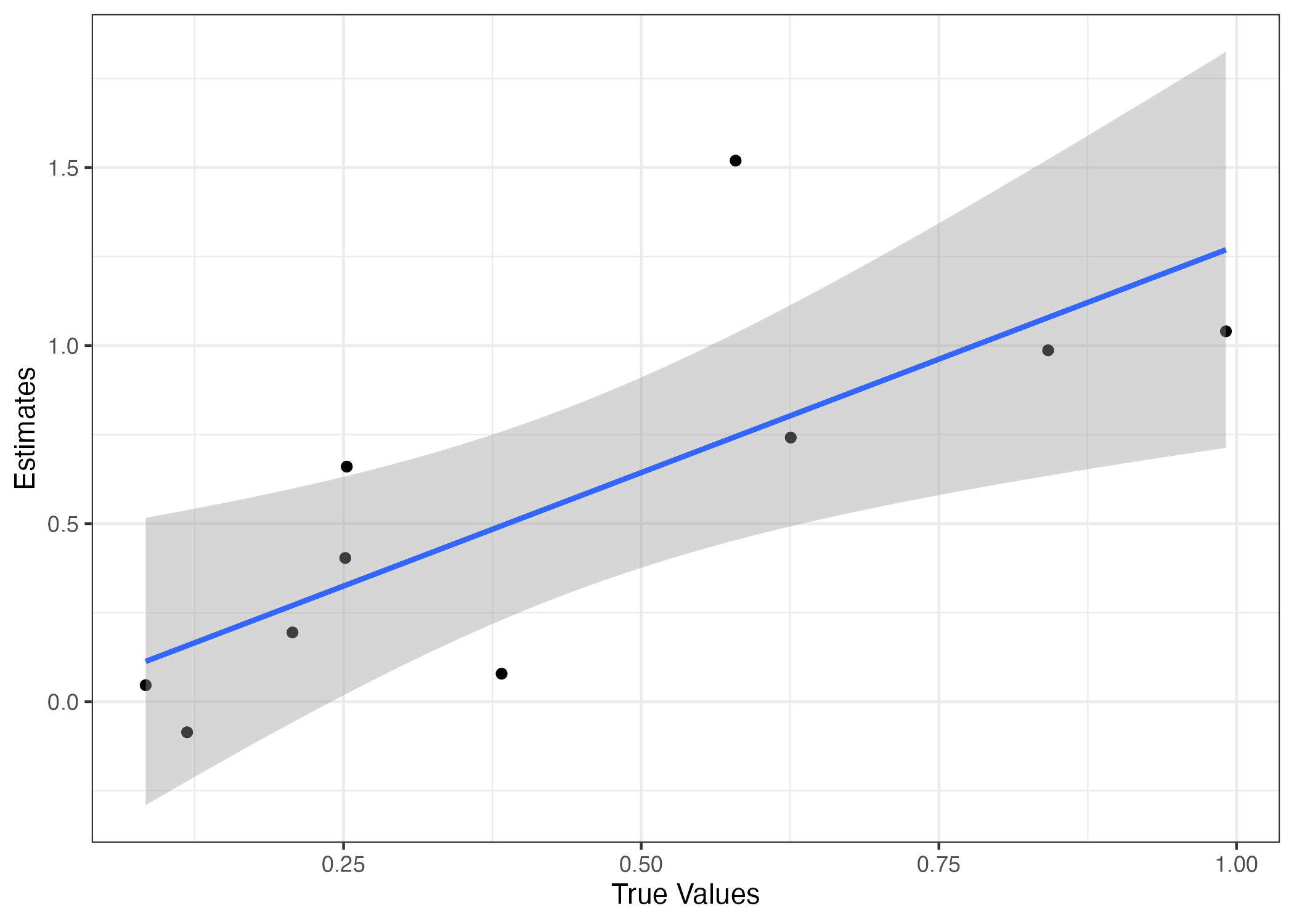}
\caption{Scatter Plot of Rule-of-Thumb Estimates}
\begin{minipage}{\linewidth}
\medskip
\footnotesize
Note: Dataset size is 5,000; reinforcement parameter is 0.99; categories with 4 or less observations dropped. Regression fit in blue; standard error bands in grey.
\end{minipage}
\label{fig9:Fig9_Drop_Estimates}
\end{figure}

Figure \ref{fig9:Fig9_Drop_Estimates} presents a scatter plot of the estimates against the true values. As indicated by the regression line, these estimates demonstrate consistency to the extent that the slope of the line approximates 1, enhancing the credibility of the analysis. However, only 10 out of 57 categories were retained in this procedure. Consequently, the results of the analysis are drastically less informative than those derived from LASSO or Ridge Regression.

Moreover, only 4 out of 57 fixed effects are significant---6 fixed effects were estimated but are nonsignificant, in addition to the 47 dropped fixed effects. In our current analysis, we study categories whose incidence probability comprises independent and conditionally independent processes as described in Section \ref{section:empirical_process}. If the presence of a dropped (infrequent) category is systematically associated with the presence of an included category, a bias may manifest in the estimates, which would then be apparent as a departure of the regression line's slope from 1. The potential for such correlations underscores a further limitation of the category aggregation approach.

\subsection{Proposed Method}

The results of the analysis conducted using our proposed model are presented in Table \ref{tab:ProposedMethodResults}. The Yule-Simon Draws and the Continuous Covariate yield estimated coefficients of 0.964 and 1.009 respectively, suggesting precise and consistent estimation given their true values of 1.

\begin{table}[tbp] 
\centering 
\caption{Regression Results using the Proposed Method} 
\label{tab:ProposedMethodResults} 
\begin{tabular}{@{\extracolsep{5pt}}lc} 
\\[-1.8ex]\hline 
\hline \\[-1.8ex] 
 & \multicolumn{1}{c}{\textit{Dependent variable:}} \\ 
\cline{2-2} 
\\[-1.8ex] & y \\ 
\hline \\[-1.8ex] 
 Yule-Simon Draws & 0.964$^{***}$ \\ 
  & (0.063) \\ 
  & \\ 
 Continuous Covariate & 1.009$^{***}$ \\ 
  & (0.014) \\ 
  & \\ 
\hline \\[-1.8ex] 
Observations & 5,000 \\ 
R$^{2}$ & 0.511 \\ 
Adjusted R$^{2}$ & 0.511 \\ 
Residual Std. Error & 1.009 (df = 4998) \\ 
F Statistic & 2,615.147$^{***}$ (df = 2; 4998) \\ 
\hline 
\hline \\[-1.8ex] 
\textit{Note:}  & \multicolumn{1}{r}{$^{*}$p$<$0.1; $^{**}$p$<$0.05; $^{***}$p$<$0.01} \\ 
\end{tabular} 
\end{table}

\begin{figure}[tbp]
\centering
\includegraphics[width=0.75\linewidth]{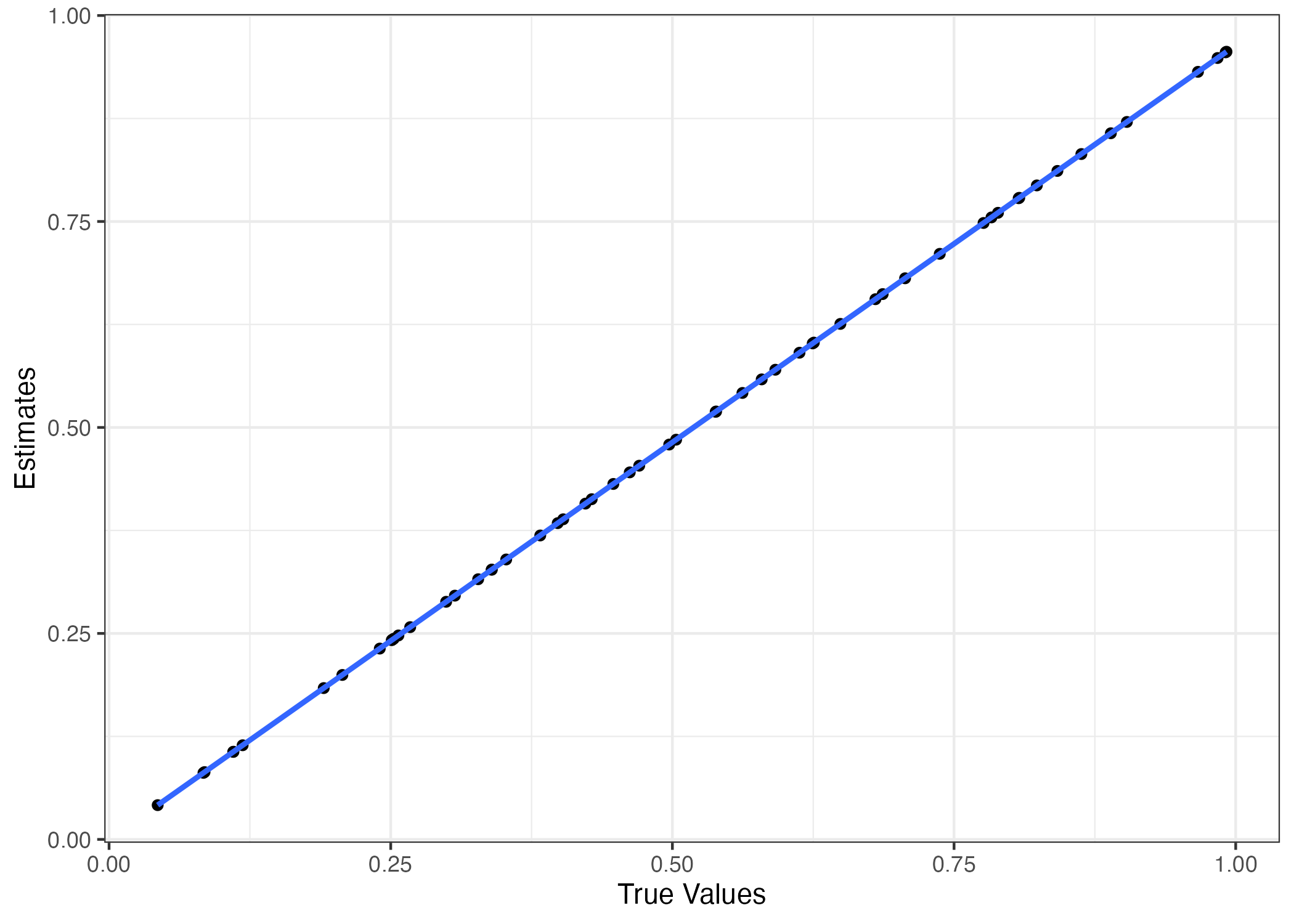}
\caption{Scatter Plot of Estimates from Proposed Procedure}
\begin{minipage}{\linewidth}
\medskip
\footnotesize
Note: Dataset size is 5,000; reinforcement parameter is 0.99. Regression fit in blue; standard error bands in grey.
\end{minipage}
\label{fig11:Fig11_Proposed_Estimates}
\end{figure}

Figure \ref{fig11:Fig11_Proposed_Estimates} presents a scatter plot of the fixed effect estimates against the true values. As both the points and the regression line visually indicate, these estimates demonstrate consistency to the extent that the slope of the line approximates 1.

The improvement in efficiency and specificity is achieved by using a map from categories to Yule-Simon draws. This technique is incorporated to simulate real-life contexts where we anticipate deploying our model. In these settings, analysts are typically aware of each category's description, such as its color, enabling a map from categories to a numerical representation. This mapping allows the creation of a higher-dimensional Hilbert space embedding for the categories, and the estimation of a finite-dimensional parameter vector sufficient to identify the category fixed effects. Through this structured approach, our model gains efficiency and maintains estimator convergence where traditional estimators fall short, as detailed earlier.

Moreover, crucially, the extent to which category fixed effects is found to be significant is decoupled from the effect size. This is because the coefficient on the Yule-Simon draw converges with increasing sample size, and the fixed effect in this model is the product of the known draw and the inferred coefficient. Therefore, categories with much smaller fixed effects are likely to be significant in this model. In essence, by introducing additional model structure, we convert the discrete estimation problem into a more efficient continuous estimation framework.

Furthermore, as the model is stable and the empirical process meets the classical conditions, the convergence of the coefficient estimates takes place across all categories and the entirety of the data. This includes both the elements that relate to a category and the elements that relate to other categories. In sum, these two features ensure that the fixed effect estimates are likely to be both consistent and precise in a wide variety of settings.

\subsection{Conclusion}

Our simulations demonstrate the limitations of applying traditional regression models to estimate complex and evolving categorical data. Specifically, we find that traditional estimators struggle to provide accurate and consistent estimates for categories that are sparse or emerging. This stems from two key issues:

First, when new categories emerge rapidly, the model becomes saturated and unable to precisely estimate the effects. Our analysis of varying reinforcement parameter values showed that higher rates of new category introduction lead to increased estimation errors.

Second, even when new categories emerge slowly, categories that are infrequently reinforced remain challenging to estimate. This manifests in diverging estimation errors between common and rare categories as the sample size grows. Our investigation into entropy and parameter credibility further supports the view that accumulating observations does not resolve the deficiencies for sparsely reinforced categories.

Critically, these characteristics are common in many real-world categorical datasets, especially those capturing evolving social constructs and classifications. However, diagnosing the limitations can be difficult. Strategies like variable selection and category aggregation may not fully address the underlying sparse data challenges.

In contrast, our proposed approach demonstrates superior consistency by substituting categorical variables with their numerical representations. This resolves the estimation difficulties by converting the problem into a more efficient continuous estimation framework. The strengths of our model include:

\begin{enumerate}
	\item Consistent and precise estimates, regardless of category dynamics.
	\item Avoidance of instability due to emerging or sparse categories.
	\item Computational scalability, as the dimensionality remains fixed.
	\item No requirement for ad-hoc solutions such as dropping categories.
\end{enumerate}

In summary, our simulations validate the advantages of our proposed approach over conventional models in estimating intricate and evolving categorical data. The method's consistency, precision, and computational efficiency position it as an attractive modern alternative to traditional regression techniques for qualitative data analysis.

\section{Empirical Analysis}

Our empirical analysis navigates the intricate relationship between color and consumer ratings in the fashion industry. Traditional economic theory posits that color should not influence evaluations of functionally identical items; yet human psychology and consumer behavior suggest a more complex reality. Colors may elicit emotional responses, convey cultural meanings, and influence personal preferences, thereby affecting product reviews and ratings.

Moreover, the complexity of inferring these relationships from data introduces further uncertainty. Whereas manufacturer and seller competition would imply that any costless factor conferring a quality advantage should be competed away, this explanation hinges on the discoverability of the factor's influence such that it can be exploited by market forces. Yet, despite the centrality of color in marketing and economics, most prior evidence has been directional in scope and confined to behavioral laboratory studies \citep{labrecque2012exciting}. Therefore, an open question remains as to the extent to which any systematic and predictable relationship exists in data from a real-world marketplace.

We examine this question using data from Amazon. Specifically, we investigate the product rating, averaged across all customer reviews for each product. Utilizing the average rating provides a consistent numerical measure of satisfaction on a 1-5 scale. Averaging accounts for individual variability and enables comparison across products. This allows us to assess whether color mentions systematically influence ratings, shedding light on underlying economic and psychological factors.

We represent the influence of color on product ratings through a mathematical model, formulated as follows:

\begin{equation}
r_i = \alpha + \gamma K(\mathrm{red}(i), \mathrm{green}(i), \mathrm{blue}(i); w) + \epsilon_i.
\end{equation}
\noindent Here, \(\alpha\) is the intercept. \(\gamma\) is the influence parameter that determines the extent to which the kernel function influences the mean rating \(r_i\) of product \(i\). \(\mathrm{red}(i), \mathrm{green}(i), \mathrm{blue}(i)\) are the RGB encodings of product color. \(\beta\) are the parameters of the kernel function \(K\). \(\epsilon_i\) is the error term. RGB values are normalized to lie within the [0, 1] interval for numerical stability, and in cases where multiple colors are referenced, the corresponding RGB values are averaged.

This model is directly comparable to the data-generating process (`Fido's Ball') presented earlier in Section \ref{sec:fido}. In particular, our specification for the average rating of each product aligns with the framework and method for estimating Fido's joy, as given in Equation \eqref{eqn:Fido3}. The mathematical representation of color in the RGB encoding parallels the selection and mapping of balls in the game of fetch between Fido and Odif, guided by the Yule-Simon draws. The versatility and multifaceted nature of the kernel functions, and their applicability to marketing and economic models, resonate with the various types of balls and their impact on Fido's joy. A direct alignment between the process analogy and empirical estimation ensures a coherent and consistent bridge between our theoretical results and empirical analysis.

A key advantage of our proposed method is its compatibility with a broad array of kernel functions. This flexibility makes our method a versatile tool for addressing challenges in the realm of categorical variables, regardless of the specific characteristics of the dataset under consideration. To make the connection between our focus on color and the broader methodology more explicit, we detail below the specific kernel functions that correspond to well-known models in marketing and economics, demonstrating how they can be applied to our analysis.

First, we consider the linear kernel, which translates to the dot product. This form of the model corresponds to a regression with the inclusion of representations from $\mathcal{Z}$ as covariates. Second, we consider the Gaussian Radial Basis Function (RBF) kernel, which translates to an ideal point model---the center point of the kernel corresponds to the idealized product (i.e., the ideal point), with deviations from the ideal point providing less utility. Third, we consider the multiquadric RBF kernel. This kernel is similar to the Gaussian RBF but distinct in the following way: whereas the Gaussian RBF peaks at the center point and diminishes exponentially with the square of the distance, the multiquadric RBF reaches its nadir at its center point, and increases at the rate of the inverse of the distance moving out from the center. We estimate these models using maximum likelihood estimation, as each corresponds to a tractable functional form---more complex kernels can be estimated using either simulated maximum likelihood or Bayesian methods such as tracing out the posterior using Monte Carlo methods (e.g., the No U-Turn Sampler).

\subsection{Results}

Table \ref{tab:kernels} details our findings from employing linear, Gaussian RBF, and multiquadric RBF kernels. Within the linear kernel model, the influence parameter is confounded with other kernel parameters due to the inherent linearity of the kernel, thus necessitating the setting of this parameter to 1 for the estimation procedure. Our results indicate a highly significant intercept, whereas the coefficient related to the Red variable is nonsignificant. The Green coefficient is significant at the 0.05 level, and the Blue coefficient is significant at the 0.01 level; however, the effect sizes are notably modest. For example, a unit increase in the Blue color value from 0 to 1 (corresponding to 0 to 255 in the original RGB model) leads to an average rating change of -0.048. Although statistically significant, this result is substantively unimportant, given that the ratings variable spans a range of 1 to 5, resulting in a maximum effect size of approximately 0.01\%.

\begin{table}[!htbp] \centering 
  \caption{Results from Linear, Gaussian RBF, and Multiquadric RBF Kernels} 
  \label{tab:kernels} 
  \begin{tabular}{
    @{\extracolsep{5pt}}
    l
    S[table-format=2.3, table-space-text-post = $^{***}$]
    S[table-format=2.3, table-space-text-post = $^{***}$]
    S[table-format=2.3, table-space-text-post = $^{***}$]
  }
  \\[-1.8ex]\hline 
  \hline \\[-1.8ex] 
  & \multicolumn{3}{c}{\textit{Kernels}} \\ 
  & \textit{Linear} & \textit{Gaussian RBF} & \textit{Multiquadric RBF} \\
  \cline{2-4} 
  Intercept & 3.838$^{***}$ & 3.698$^{***}$ & 4.221$^{***}$ \\
  Red & -0.013 & 0.442$^{***}$ & 0.427$^{***}$ \\
  Green & 0.039$^{**}$ & 0.537$^{***}$ & 0.552$^{***}$ \\
  Blue & -0.048$^{***}$ & 0.276$^{***}$ & 0.297$^{***}$ \\
  Influence & {} & 0.211$^{***}$ & -0.319$^{***}$ \\
  \hline \\[-1.8ex] 
  Observations & {105,272} & {105,272} & {105,272} \\ 
  Log-likelihood & {-179329} & {-179281} & {-179278} \\
  \hline 
  \hline \\[-1.8ex] 
  \textit{Note:}  & \multicolumn{3}{r}{$^{*}$p$<$0.1; $^{**}$p$<$0.05; $^{***}$p$<$0.01} \\ 
  \end{tabular} 
\end{table}

The influence function estimates within the Gaussian RBF and multiquadric RBF models underscore the critical role of a kernel capable of projecting the initial RGB embedding into a significantly higher-dimensional feature space. As detailed in Table \ref{tab:kernels}, the estimates for this parameter are 0.211 and -0.319 in the Gaussian RBF and multiquadric RBF model specifications, respectively. This divergence translates to contrasting behaviors in the kernels: In the Gaussian RBF, the kernel value is 1 at the center point and decreases with the Euclidean distance from this point, as evidenced by the exponential of the negative square of the distance. Conversely, the multiquadric RBF begins with a value of 1 at the center point but increases as the distance from the center point expands. Consequently, the findings in the Gaussian RBF model, where the ratings decrease moving away from the center point (indicated by the positive influence coefficient), are mirrored by the negative influence coefficient of the multiquadric RBF, where the kernel's value augments with increasing distance.

In both the Gaussian RBF and multiquadric RBF models, the center point bears the same substantive interpretation: it signifies the color where the average rating peaks. Specifically, our analysis identifies this color as [113, 137, 70] in the Gaussian RBF model and [109, 141, 76] in the multiquadric RBF model. Colloquially referred to as a shade of dirty olive green, and formally aligned closest to `Dingley'---an `unsaturated light cold chartreuse,' this color emerges as the most favored in Amazon fashion. Its popularity can likely be attributed to its versatility across genders, various garments, and occasions, making it a standout choice in our dataset.

Critically, our findings demonstrate the distinct superiority of the two RBF kernel models over the linear kernel, underscoring the importance of allowing for a rich and expansive feature space derived from the RGB representation of color references. Among the two RBF kernels, the multiquadric kernel marginally outperforms the Gaussian RBF in representation, as evidenced by the log-likelihood. This suggests that the decline in expected ratings with distance from the center point manifests more gradually than an exponential decrease and is more aptly characterized by the multiquadric function. Thus, the analysis not only showcases the value of employing nonlinear kernel models but also highlights the nuanced differences between them in accurately capturing the underlying dynamics of color preferences.

More broadly, these results highlight a consistent yet hitherto unexploited influence of color on customer ratings within a highly competitive market of over 100,000 products from various sellers. The fact that such a systematic and enduring impact has not been competed away may indicate that market participants using extant estimators have been unable to decipher this relationship. That is, the data is consistent with the explanation that traditional estimators falter in uncovering this intricate link between a seemingly minor qualitative variable like color and significant market outcomes. Our proposed model seeks to circumvent these limitations by employing additional structure relating to the categories described in the qualitative data, thereby enhancing efficiency and ensuring consistency.

\section{General Discussion}

We introduce a novel framework for integrating qualitative data into quantitative models for causal estimation. This methodology employs functional analysis, embedding observed categories into a latent Baire space, and mapping them through a continuous linear map—a Hilbert space embedding—to an RKHS of representations. By leveraging the kernel trick and transfer learning, we streamline the estimation process. Validation through extensive simulations and a case study shows that our model outperforms traditional methods, particularly when managing complex and dynamic qualitative information.

We develop our paper in the context of unstructured qualitative data, where challenges such as category dynamics and sparsity are particularly pronounced. However, the underlying mathematical theory and model structure have broader applicability and can be adapted even when categorical variables are directly observed. For example, in a demand model for automobile sales, a car's color might be treated as a categorical variable, given its influence on consumers' choices between different colors. Our model can be applied using the RGB representation of color, paralleling how color might be expressed in unstructured textual data. This adaptability to different data types highlights the potential for our modeling proposals to extend well beyond the confines of unstructured data.

Our research reveals a distinct dichotomy between intuition and mathematical admissibility. Specifically, while some admissible functions in accordance with Mercer's theorem may seem elementary, such as the dot product, other intuitive choices, like the Euclidean norm, contradict Mercer's condition, leading to inconsistency with the inner product in a feature space. This observation underscores the critical role played by the development of a framework establishing clear requirements that correspond to desired properties. Moreover, as the model is compatible with any positive semi-definite kernel, it offers many options for applied researchers to discover the best fitting model for a particular application. Thus, for instance, an analyst may choose from a variety of radial basis functions that both meet Mercer's condition and capture the influence of the Euclidean norm on an outcome of interest.

A limitation of our model is that, although it efficiently handles related categories that can be condensed into a singular schema, it is less effective when dealing with many unrelated or orthogonal categories. Moreover, potential challenges with feature sparsity and the extraction of continuous covariates from qualitative data also point to future model extensions. Specifically, it would be interesting to extend the model to account for feature sparsity in qualitative data, such as through the use of the Representer theorem. In addition, while we can easily extract and map simple variables like price or size using modern tools from qualitative data, intricate information (such as the emotional content in a user review) might require further model adaptations.

We hope these ideas further the use of functional analysis in incorporating qualitative data into causal econometric models. The framework we present here offers a modern approach to harnessing unstructured data that holds significant promise.

% Acknowledgements and Disclosure of Funding should go at the end, before appendices and references

\acks{This research was supported by the Ministry of Education (MOE), Singapore, under its Academic Research Fund (AcRF) Tier 2 Grant, No. MOE-T2EP40221-0008.}


\newpage

\bibliography{References}

\end{document}